%% file: main.tex
\PassOptionsToPackage{table}{xcolor}
\documentclass[10pt,twocolumn]{article}
\ifdefined\pdfminorversion\pdfminorversion=7\fi
\usepackage[preprint]{antintlpaper}
\usepackage{xurl}
\usepackage[utf8]{inputenc}
\usepackage{multirow,adjustbox}
\usepackage{tikz,pifont}
\setcitestyle{authoryear,open={[},close={]}}

\newcommand{\stitle}[1]{\vspace{1mm} \noindent {\bf #1}}
\newcommand{\method}{\textsc{WorldIDE}}

\definecolor{GroupGray}{RGB}{238,238,238}
\definecolor{MethodBlue}{RGB}{226,239,255}
\definecolor{CaseGreen}{RGB}{46,125,50}
\definecolor{CaseRed}{RGB}{198,40,40}
\newcolumntype{Y}{>{\raggedright\arraybackslash}X}
\newcolumntype{R}{>{\raggedleft\arraybackslash}X}

\input{numbers}

\AntTitle{Supervise Relations, Not Answers: Relational State Identification for Spatial World Models}
\AntRunningTitle{Supervise Relations, Not Answers}
\AntAuthors{%
  Yihao~Hu\texorpdfstring{\textsuperscript{1,2,*}}{}\and
  Pan~Wang\texorpdfstring{\textsuperscript{1,3,*}}{}\and
  Yitong~Lin\texorpdfstring{\textsuperscript{1,*}}{}\and
  Xianglong~Zeng\texorpdfstring{\textsuperscript{3}}{}\and
  Deyuan~Liu\texorpdfstring{\textsuperscript{2}}{}\and
  Mingda~Wan\texorpdfstring{\textsuperscript{2}}{}\and
  Hang~Wang\texorpdfstring{\textsuperscript{1}}{}\and
  Yongqi~Tong\texorpdfstring{\textsuperscript{1}}{}\and
  Wei~Wu\texorpdfstring{\textsuperscript{1,\textdagger}}{}\and
  Zhihao~Wen\texorpdfstring{\textsuperscript{1,\textdagger}}{}}
\AntAffiliations{%
  \textsuperscript{1}Ant International \quad
  \textsuperscript{2}Westlake University \quad
  \textsuperscript{3}University of Science and Technology of China}
\AntContact{\textsuperscript{*}Equal contribution. \quad
  \textsuperscript{\textdagger}Corresponding authors: Wei Wu and Zhihao Wen}
\AntDate{}
\AntAbstract{
Spatial answers do not uniquely determine the world they describe: two visual
histories can agree on when a sink and a plant appeared together but not on
which way the observer turned. Trained on one answer at a time, a model can
therefore merge histories that other questions must tell apart, which we call
\emph{query-induced state aliasing}. We instead cast spatial representation
learning as \emph{relational state identification}: a representation should
keep every distinction that a family of grounded operations needs. \method{},
an 8B vision-language model fine-tuned on four complementary relation families
(categorical, metric, trajectory, and observer motion) with source-verifiable
targets, improves mean accuracy on OST-Bench, UCS-Bench, and SIS-Bench by
13.25 points over its matched base model. Controlled history pairs show that
\method{} better separates histories that differ in timing, and geometric
probes find that models with similar accuracy can recover spatial quantities
with different consistency. In short, relations teach what single answers
leave out, and probes reveal what accuracy alone hides.
}

\begin{document}
\raggedbottom
\makeanttitle

\input{sections/introduction}

\input{sections/motivation}

\input{sections/method}
\input{sections/experiments}
\input{sections/related_work}
\input{sections/conclusion}
\input{sections/limitations}

\bibliographystyle{plainnat}
\bibliography{references}

\clearpage
\appendix
\makeatletter
\@addtoreset{table}{section}
\@addtoreset{equation}{section}
\makeatother
\renewcommand{\thetable}{\thesection\arabic{table}}
\renewcommand{\theHtable}{\thesection\arabic{table}}
\renewcommand{\theequation}{\thesection\arabic{equation}}
\renewcommand{\theHequation}{\thesection\arabic{equation}}
\input{sections/appendix}

\end{document}

%% file: numbers.tex
\newcommand{\BaseOST}{50.78}
\newcommand{\WorldIDEOST}{64.55}
\newcommand{\DeltaOST}{+13.78}
\newcommand{\BaseUCS}{45.24}
\newcommand{\WorldIDEUCS}{58.39}
\newcommand{\DeltaUCS}{+13.15}
\newcommand{\BaseSIS}{65.09}
\newcommand{\WorldIDESIS}{77.90}
\newcommand{\DeltaSIS}{+12.81}
\newcommand{\BaseMean}{53.70}
\newcommand{\WorldIDEMean}{66.95}
\newcommand{\DeltaMean}{+13.25}

\newcommand{\JointMean}{66.86}

\newcommand{\WorldIDEShared}{77.90}
\newcommand{\WorldIDEDeranged}{52.16}
\newcommand{\SharedEncodeReduction}{67.03}

%% file: sections/introduction.tex
\section{Introduction}

An embodied observer sees a sink and a plant and later changes its heading.
Determining when the objects appeared together requires their observation
history, whereas recovering the observer's turn requires a reference frame.
These questions concern the same causal episode but depend on different
aspects of its history (Figure~\ref{fig:teaser}). This motivates our central
question: \emph{which distinctions between histories can relational
supervision help preserve for different spatial and temporal operations?}

Spatially grounded models learn geometric relations from visual observations
\citep{chen2024spatialvlm,cheng2024spatialrgpt,wang2026ucs}, streaming video
systems retain evidence through visual memories
\citep{song2024moviechat,zhang2025flashvstream,zeng2026onestreamer}, and world
models learn dynamics for prediction and planning
\citep{hafner2025dreamerv3,bruce2024genie,assran2025vjepa2,chen2026codeworldmodel}.
Spatial supervision commonly pairs a question, its evidence, and a target
answer.

\input{figures/teaser}

Learning operation-relative spatial representations from causal video raises
three challenges. \textbf{(C1)}~The first is to specify which distinctions
individual answers leave unresolved. Two histories can show the sink and the
plant together at the same moment yet differ in the observer's later turn;
supervising only the co-occurrence answer lets a representation merge them, a
potential failure mode we call \emph{query-induced state aliasing}.
\textbf{(C2)}~The second is to supervise these distinctions verifiably. No
single relation fixes them all: left--right arrangement leaves distance free,
and co-occurrence leaves the observer's turn free. Relations over geometry,
trajectories, and observer motion must therefore be combined, with targets
checkable against the source rather than predicted by a model that may share
the aliasing. \textbf{(C3)}~The third is to evaluate what a representation
preserves beyond aggregate accuracy, which scores each question on its own and
cannot show whether models with similar scores keep the same distinctions.
This requires testing the representation itself: whether paired histories stay
distinguishable, whether one encoding is reusable yet bound to its source
video, and whether geometry is explicitly recoverable.

Motivated by predictive state representations \citep{singh2004psr}, we
characterize spatial state by its recoverable relations.
\emph{Relational state identification} treats complementary operations as
constraints on which distinctions a representation should preserve (C1).
\method{} instantiates this principle with four complementary operation
families: categorical relations, metric geometry, object trajectories, and
observer motion. Source geometry and released annotations provide their
targets (C2; Figure~\ref{fig:overall}a). We assess which distinctions remain
accessible through paired histories, repeated readout of a shared
representation, and explicit geometric probing
(C3; Figure~\ref{fig:overall}b,c).
Across three complete spatial reasoning benchmarks
\citep{lin2025ost,wang2026ucs,zou2026sis}, \method{} improves performance
under matched evaluation. Beyond aggregate accuracy, paired-history
comparisons and matched geometric probes connect relational training with two
requirements of spatial state: distinguishing histories that demand different
answers and recovering compatible relationships within a history.
Our contributions are:
\begin{itemize}[leftmargin=*,topsep=-1pt,itemsep=1.5pt,parsep=0pt,partopsep=0pt]
  \item We formulate \emph{relational state identification}, characterize
  operation-relative state sufficiency (Proposition~1), and name
  \emph{query-induced state aliasing} as a potential failure mode of
  individual-answer supervision.
  \item We introduce \method{}: four complementary, source-verifiable
  relation families over causal video, with functional evaluations of
  repeated readout, source dependence, and explicit spatial recoverability.
  \item We identify operation-dependent effects of trajectory supervision
  on history distinguishability, and geometric differences between models
  with similar accuracy.
\end{itemize}

%% file: figures/teaser.tex
\begin{figure}[!t]
\centering
\includegraphics[width=1.0\linewidth]{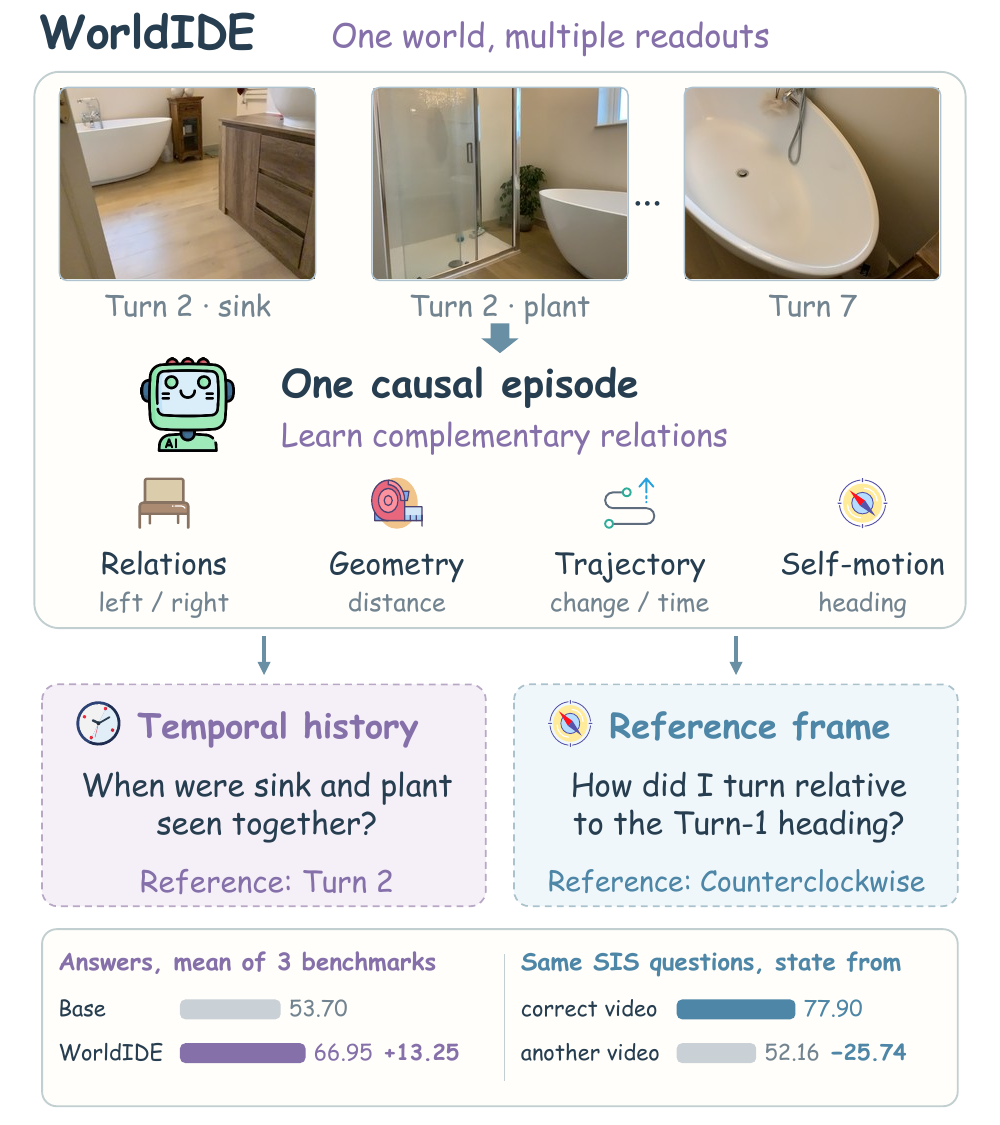}
\vspace{-4mm}
\caption{\textbf{One causal episode, multiple relational readouts.} One OST-Bench
episode (top) illustrates four possible operation targets (middle), rather
than a jointly supervised training record. Reference answers use each query's
own causal endpoint: Turn~5 for co-occurrence, Turn~7 for rotation.
Bottom: the three-benchmark mean rises from 53.70 to 66.95
(Table~\ref{tab:main-results}), and another video's state drops SIS from
77.90 to 52.16 (Table~\ref{tab:functional-tests}b).}
\label{fig:teaser}
\vspace{-4mm}
\end{figure}

%% file: sections/motivation.tex
\section{Problem: Relational State Identification}
\label{sec:problem}
\label{sec:motivation}

\stitle{Individual answers alias world states.}
Let $h=o_{\leq t}$ be a causal visual history and $\mathcal R$ a family of
grounded operations. Each $r\in\mathcal R$ fixes an operator, entity
arguments, a reference frame, and a temporal scope, and has a source-grounded
target $y_r(h)$. The \emph{relation signature}
$\rho_{\mathcal R}(h)=(y_r(h))_{r\in\mathcal R}$ induces an equivalence
$h\sim_{\mathcal R}h'$ when the signatures agree, and the operation-relative
state of $h$ is its class $[h]_{\sim_{\mathcal R}}$. Single-operation
supervision permits $E(h)=E(h')$ when $y_{r_a}(h)=y_{r_a}(h')$ but
$y_{r_b}(h)\neq y_{r_b}(h')$. No deterministic reader of encoder $E$ can
then answer $r_b$ correctly on both histories; this is \emph{query-induced
state aliasing}. Appendix~\ref{app:theory} shows that a query-independent encoder admits
readers for every operation in $\mathcal R$ if and only if it never merges
histories with different signatures (Proposition~1).

\stitle{Functional state abstraction.}
Histories remain equivalent despite visual or unqueried differences when
all targets agree. This state preserves operation-relevant distinctions
without reconstructing the complete physical scene.

\stitle{Complementary partitions of history.}
Two histories can place an object at equal distances on opposite sides of
the observer: distance groups them together, while signed bearing separates
them. Identical final geometry can also conceal different first-seen times.
An operation family's state is defined by the intersection of these
answer-induced partitions. Complementary targets encourage these spatial
and temporal distinctions.

\stitle{Testing operation-relative distinctions.}
Pairs in Eq.~\ref{eq:app-cross-operation-ambiguity} agree under $r_a$ but
differ under $r_b$. PairAcc counts pairs with both $r_b$ predictions correct,
measuring behavioral distinctions unresolved by $r_a$ without directly
establishing latent representation collisions.

\stitle{Learning relative to an operation family.}
WorldIDE distributes these constraints across training histories and
evaluates shared-state sufficiency empirically. Enlarging $\mathcal R$
refines the target equivalence classes
(Eq.~\ref{eq:app-refinement}).

%% file: sections/method.tex
\section{Method: \method{}}
\label{sec:method}

Figure~\ref{fig:overall} connects relational training to native benchmark
readout and auxiliary State Plane probes of spatial recoverability from
frozen features.

\input{figures/overall}

\subsection{Source-Grounded Relational Supervision}
\label{sec:relational-learning}

\stitle{Supervising relational information.}
Spatial answer supervision associates a question $q$ about history $h$ with
an annotated answer $a$ \citep{NEURIPS2023_6dcf277e,chen2024spatialvlm}:
\begin{equation}
  \mathcal L_{\mathrm{QA}}(\theta)
  =
  \mathbb E_{(h,q,a)\sim\mathcal D_{\mathrm{QA}}}
  \left[-\log p_\theta(a\mid h,q)\right].
  \label{eq:qa-objective}
\end{equation}
\method{} trains on operation-conditioned, source-grounded targets $y_r(h)$:
\begin{equation}
  \mathcal L_{\mathrm{rel}}(\theta)
  =
  \sum_{f\in\mathcal F}\alpha_f\,
  \mathbb E_{\mathcal D_f}
  \bigl[-\log p_\theta(y_r(h)\mid h,r)\bigr].
  \label{eq:relational-objective}
\end{equation}
Targets come from source geometry or released annotations
(Appendix~\ref{app:relational-training}), without exposing geometry as visual
input. This source-grounding principle is related to decision-grounded
supervision for tool-use agents, where training actions are supported by the
learner-visible interaction history \citep{hu2026trajectory2gym}.
Natural-language requests $r$ specify entities, reference frame, and
temporal scope. Family loss masses $\alpha_f$ allocate emphasis independently
of record count. Both objectives predict answer tokens; WorldIDE specifies
target provenance, operation semantics, and complementary coverage.

Multiple distinct requests occur for 2.61\% of normalized visual prefixes;
this measures textual request coverage rather than distinct semantic
relations. Complementarity is primarily learned across the training
distribution. Target construction and family allocation appear in
Appendix~\ref{app:relational-training}.

\stitle{Complementary spatial constraints.}
Categorical relations supervise qualitative arrangements, while metric
geometry targets distance, displacement, and signed direction. Object
trajectories constrain spatial changes across observations; observer motion
links camera movement to reference-frame changes.

\stitle{Temporal and reference-frame dependencies.}
Camera rotation changes a stationary object's observer-relative direction;
histories with identical final configurations can also differ in first-seen
time. Observer-motion supervision constrains descriptions across reference
frames, while trajectory supervision connects present observations with
earlier spatial events. These operations require different information
about causal histories.

\stitle{Grounded operation examples.}
An archived metric request fixes the entity and frame: camera-to-TV
distance at frame 27 of 32, with source-derived nearest-point target
1.4\,m. A temporal request asks for the first-appearance order of blanket,
door, and couch, with target ``door, couch, blanket.'' Both use the same
prediction objective; paired-history tests assess the resulting distinctions.

\subsection{Query-Independent State Reuse}
\label{sec:state-materialization}

A reusable spatial representation should support multiple questions about
the same causal history without depending on any particular question.
We examine this property through the observation-conditioned visual prefix.
Let $p(h)$ contain the causal observations and question-independent context,
ending before question-dependent tokens. Its representation is
\begin{equation}
  Z_h=\operatorname{Prefill}_{\theta}(p(h)),
  \label{eq:visual-state}
\end{equation}
and the same language reader accesses it for each question $q_j$:
\begin{equation}
  \widehat y_j=D_\theta(q_j;Z_h),
  \qquad j=1,\ldots,m.
  \label{eq:shared-readout}
\end{equation}
Independent decoding contexts allow each question to read the same $Z_h$.
Standard prefix caching implements this shared access
\citep{Kwon_2023,NEURIPS2024_724be447,MLSYS2024_a66caa17,di2025streaming};
Section~\ref{sec:rq2} tests repeated readout and source dependence.

\stitle{Source dependence.}
To measure the contribution of the observed history, a fixed, answer-blind
permutation $\pi$ substitutes another video's state:
\begin{equation}
  \widehat y_j^{\,\mathrm{der}}
  =
  D_\theta(q_j;Z_{\pi(h)}),
  \label{eq:source-derangement}
\end{equation}
with question and reader fixed. This contrast measures source dependence;
Appendix~\ref{app:shared-state} specifies the matched execution protocol.

\subsection{Executable Spatial Readout}
\label{sec:executable-readout}

\stitle{Entity-conditioned fields from shared features.}
The endpoint features $H_h$ are constructed independently of subsequent
queries. For each eligible entity $e$, the matched auxiliary head recovers
\begin{equation}
  \widehat S_{h,e}=G_\phi(H_h,e).
  \label{eq:state-plane}
\end{equation}
The endpoint collection is
\begin{equation}
  \widehat S_h=\{\widehat S_{h,e}\}_{e\in\mathcal E_h}.
  \label{eq:entity-state-collection}
\end{equation}
All queries share $H_h$. Source-provided, unambiguous entity labels specify
the referents; coordinates, bearings, and event times remain prediction
targets. The recovered fields are conditioned on entity identity but
independent of the subsequent operation. The probe therefore measures
geometric and temporal recoverability given grounded entities.

Following probing practice
\citep{alain2018understandingintermediatelayersusing,hewitt-liang-2019-designing,belinkov-2022-probing},
each frozen encoder receives a separately fitted probe under matched settings.
The learned readout measures accessibility under that protocol. Entity
selection, probe architecture, and the earlier single-track protocol appear
in Appendix~\ref{app:state-plane}.

\stitle{Field-level executable semantics.}
A grounded operation reads the recovered fields using the entity, reference
frame, and temporal scope specified by its request:
\begin{equation}
  \widetilde y_r
  =
  \operatorname{Exec}(r,\widehat S_h).
  \label{eq:typed-execution}
\end{equation}
From one endpoint, entity-conditioned fields support threshold, sector,
visibility, and temporal queries; pair operations combine two entities' fields.
The requested operation selects which fields the fixed executor uses
(Appendix~\ref{app:taskblind-plane}).
Rotating the observer heading by $\varphi$ shifts bearing by $-\varphi$ and
leaves distance unchanged. The archived reference-frame tests verify these
transformation rules; source-derived targets separately assess prediction
accuracy (Appendix~\ref{app:state-plane}). Distance, displacement,
and angle are predicted independently, so accuracy and cross-field
consistency are scored separately (Table~\ref{tab:state-recoverability}).

\stitle{Cross-field geometric compatibility.}
For entity $e$ at endpoint $h$, we measure agreement between independently
predicted distance and displacement fields:
\begin{equation}
  r_d(h,e)=
  \left|
    \widehat d_{h,e}-
    \sqrt{\widehat{\Delta x}_{h,e}^{\,2}
         +\widehat{\Delta z}_{h,e}^{\,2}}
  \right|.
  \label{eq:distance-compatibility}
\end{equation}
All three fields refer to the same source-annotated centroid in the endpoint
camera's right--forward plane, measured in meters. Their reference distance
is the planar displacement norm, rather than a three-dimensional or
nearest-surface distance. The residual tests a geometric identity among
independent predictions; source-target errors assess geometric correctness.
Appendix~\ref{app:state-joint-final} specifies aggregation.

%% file: figures/overall.tex
\begin{figure*}[t]
\centering
\includegraphics[width=0.96\textwidth]{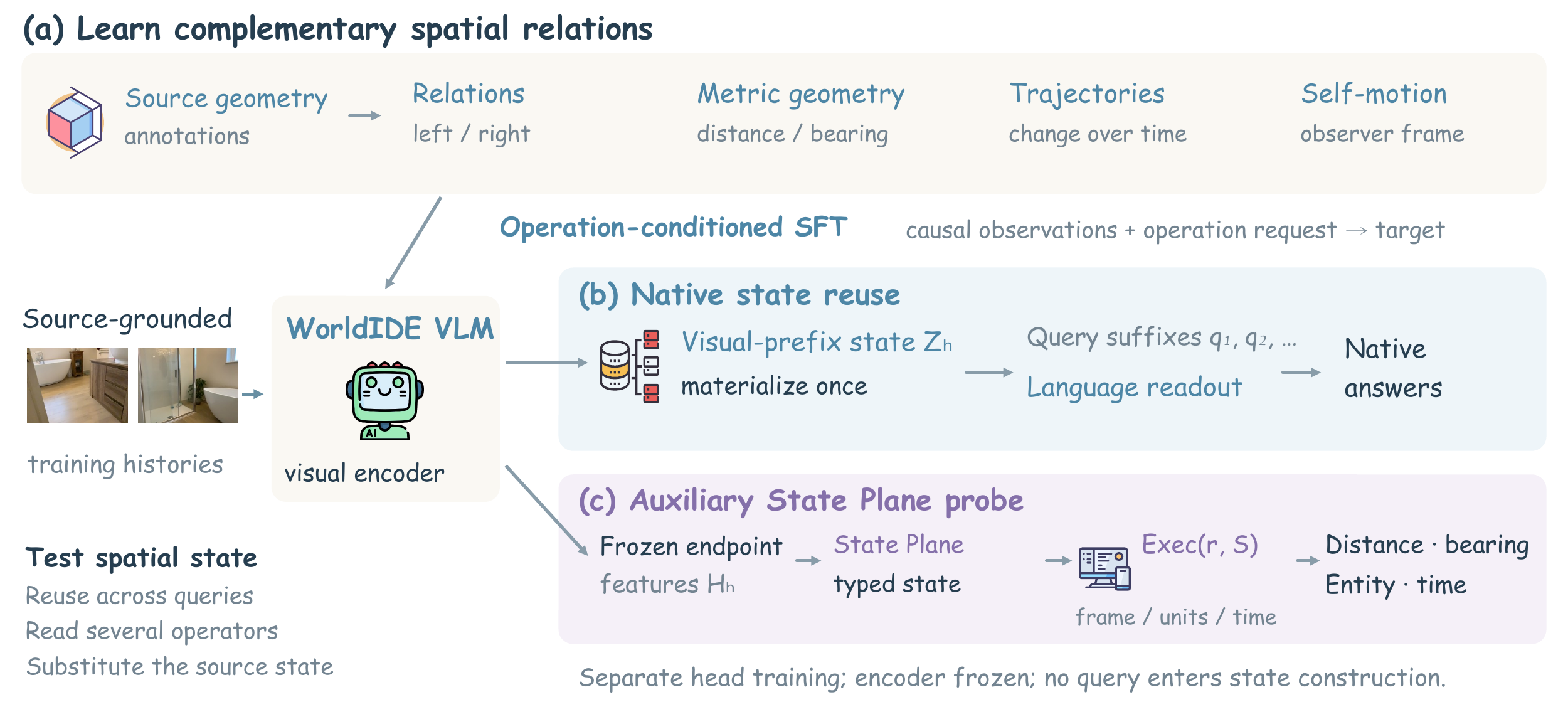}
\vspace{-2mm}
\caption{\textbf{Train on relations once, then test the state two ways.}
(a) Training: four relation families with targets computed from source geometry
or supplied by released annotations train one visual-language model
(Eq.~\ref{eq:relational-objective}). The families are organized across the
training distribution; an individual history need not carry all four.
(b) Native readout: the visual prefix of
a history is materialized once as $Z_h$ and reused by every question;
swapping in another video's $Z_h$ is the source-derangement test (RQ2).
(c) Typed readout: a separately trained head maps frozen endpoint features
$H_h$ to typed fields $\widehat S_h$ ($S$ in the panel) that deterministic
operators execute (RQ3). Benchmark answers use (b) only; (c) is a probe.}
\label{fig:overall}
\vspace{-4mm}
\end{figure*}

%% file: sections/experiments.tex
\section{Experiments and Analysis}
\label{sec:experiments}

We evaluate spatial reasoning and cross-operation distinguishability (RQ1),
repeated source-dependent readout (RQ2), and geometric recoverability under
matched probes (RQ3).

\begin{table}[b]
\caption{\textbf{Performance across three spatial benchmarks} (\%). Ref.: strongest
published model in Tables~\ref{tab:ost-main}--\ref{tab:sis-main} (own
protocol); Base and \method{}: matched evaluations (bold:
better); equal-weight mean.}
\label{tab:main-results}
\centering
\small
\setlength{\tabcolsep}{3.2pt}
\resizebox{\columnwidth}{!}{%
\begin{tabular}{llrrrr}
\toprule
Benchmark & Strongest listed ref. & Ref. & Base & \method{} & $\Delta$ \\
\midrule
OST & Gemini-2.0-Flash (Think.) & 54.2 & \BaseOST{} & \textbf{\WorldIDEOST{}} & \DeltaOST{} \\
UCS & DirectMe & 50.5 & \BaseUCS{} & \textbf{\WorldIDEUCS{}} & \DeltaUCS{} \\
SIS & Gemini-3-Flash & 71.6 & \BaseSIS{} & \textbf{\WorldIDESIS{}} & \DeltaSIS{} \\
\midrule
Mean & -- & -- & \BaseMean{} & \textbf{\WorldIDEMean{}} & \DeltaMean{} \\
\bottomrule
\end{tabular}}
\end{table}

\begin{table*}[t]
\caption{\textbf{Fine-grained results on OST-Bench} (\%). Published rows use
their reported protocols; the final block is our matched evaluation (bold:
better row). JUD./TEMP./CNT./EST.: judgement, temporal localization,
counting, estimation.}
\label{tab:ost-main}
\centering
\fontsize{7}{8}\selectfont
\setlength{\tabcolsep}{1.9pt}
\renewcommand{\arraystretch}{0.96}
\begin{adjustbox}{max width=\textwidth}
\begin{tabular}{lcccccccccccccccc}
\toprule
& \multirow{3}{*}{Avg}
& \multicolumn{4}{c}{Agent State}
& \multicolumn{5}{c}{Agent Visible Information}
& \multicolumn{6}{c}{Agent--Object Spatial Relationship}\\
\cmidrule(lr){3-6}\cmidrule(lr){7-11}\cmidrule(lr){12-17}
& & \multicolumn{2}{c}{Position} & \multicolumn{2}{c}{Orientation}
& \multicolumn{2}{c}{Existence} & Quantity & Diversity & Order
& \multicolumn{3}{c}{Direction} & \multicolumn{3}{c}{Distance}\\
\cmidrule(lr){3-4}\cmidrule(lr){5-6}\cmidrule(lr){7-8}
\cmidrule(lr){9-9}\cmidrule(lr){10-10}\cmidrule(lr){11-11}
\cmidrule(lr){12-14}\cmidrule(lr){15-17}
Model & & JUD. & EST. & JUD. & EST. & JUD. & TEMP. & CNT. & JUD. & JUD.
& JUD. & TEMP. & EST. & JUD. & TEMP. & EST.\\
\midrule
\rowcolor{GroupGray}\multicolumn{17}{l}{\textit{Representative published MLLMs}}\\
Claude-3.5-Sonnet & 47.8 & 65.1 & 36.2 & 50.6 & 30.3 & 85.8 & 67.0 & 57.9 & 57.1 & 60.0 & 39.2 & 18.3 & 21.9 & 43.8 & 54.9 & 19.0\\
Gemini-2.0-Flash & 49.5 & 59.7 & 27.3 & 56.7 & 36.5 & 89.6 & 70.8 & 59.4 & 78.7 & 55.6 & 42.0 & 17.8 & 21.5 & 45.5 & 48.9 & 27.1\\
Gemini-2.0-Flash (Thinking) & 54.2 & 57.4 & 36.3 & 61.1 & 33.4 & 88.1 & 74.8 & 61.9 & 63.6 & 73.1 & 50.9 & 51.7 & 23.3 & 51.1 & 56.8 & 22.7\\
GPT-4.1 & 53.4 & 64.2 & 30.7 & 60.8 & 33.2 & 90.8 & 78.0 & 60.6 & 82.1 & 70.8 & 51.5 & 23.4 & 28.6 & 44.9 & 53.6 & 23.9\\
GPT-4o & 48.7 & 55.6 & 20.5 & 45.6 & 33.6 & 90.6 & 75.6 & 59.8 & 78.2 & 59.6 & 46.1 & 19.5 & 21.4 & 43.1 & 50.6 & 20.4\\
InternVL-2.5-38B & 50.8 & 56.7 & 31.8 & 54.6 & 38.4 & 91.7 & 74.7 & 61.1 & 79.8 & 62.1 & 42.1 & 20.6 & 27.7 & 42.7 & 42.5 & 28.1\\
InternVL-2.5-78B & 51.1 & 60.8 & 34.4 & 49.9 & 40.7 & 90.7 & 74.4 & 65.9 & 77.9 & 61.2 & 43.4 & 22.4 & 17.8 & 46.7 & 44.4 & 22.9\\
Qwen2.5-VL-32B & 46.9 & 51.0 & 31.1 & 53.5 & 39.4 & 85.3 & 64.8 & 59.2 & 73.4 & 41.8 & 39.5 & 24.9 & 25.7 & 43.6 & 39.1 & 20.3\\
\midrule
\rowcolor{GroupGray}\multicolumn{17}{l}{\textit{Our complete Qwen3-VL-8B evaluation}}\\
Official base & 50.78 & 57.56 & 41.23 & 55.77 & 37.23 & 84.30 & 65.66 & 61.27 & 78.28 & 62.95 & 44.19 & 25.80 & 19.01 & 46.91 & 44.03 & 36.49\\
\rowcolor{MethodBlue}\textbf{\method{}} & \textbf{64.55} & \textbf{65.68} & \textbf{47.81} & \textbf{63.31} & \textbf{43.69} & \textbf{94.69} & \textbf{84.07} & \textbf{79.05} & \textbf{89.89} & \textbf{82.05} & \textbf{61.77} & \textbf{41.77} & \textbf{44.17} & \textbf{59.20} & \textbf{61.84} & \textbf{39.60}\\
$\Delta$ & +13.78 & +8.12 & +6.58 & +7.55 & +6.46 & +10.39 & +18.41 & +17.78 & +11.61 & +19.09 & +17.58 & +15.97 & +25.17 & +12.29 & +17.82 & +3.12\\
\bottomrule
\end{tabular}
\end{adjustbox}
\end{table*}

\begin{table*}[t]
\caption{\textbf{Complete UCS-Bench results} over 8,114 questions (\%): each
group's aggregate (All) and two subcategories. Published rows use their
reported protocols; final block: bold is the better matched row, $\Delta$
is \method{} minus base.}
\label{tab:ucs-main}
\centering
\fontsize{7}{8}\selectfont
\setlength{\tabcolsep}{3.6pt}
\renewcommand{\arraystretch}{0.96}
\begin{adjustbox}{max width=\textwidth}
\begin{tabular}{lcc*{13}{c}}
\toprule
& & & \multicolumn{3}{c}{Trajectory} & \multicolumn{3}{c}{Position}
& \multicolumn{3}{c}{Proximity} & \multicolumn{3}{c}{Category} & Overall\\
\cmidrule(lr){4-6}\cmidrule(lr){7-9}\cmidrule(lr){10-12}\cmidrule(lr){13-15}
Model & Size & Frames & All & My & Obj. & All & Ego & Rel. & All & Dist. & Reach & All & Cat. & Qty. &\\
\midrule
\rowcolor{GroupGray}\multicolumn{16}{l}{\textit{Representative official references}}\\
DeepSeek-V3 & -- & 50 & 42.0 & 41.0 & 44.8 & 33.2 & 35.4 & 31.6 & 37.7 & 45.9 & 31.2 & 32.3 & 36.8 & 30.3 & 35.6\\
GPT-5 & -- & 50 & 44.5 & 43.3 & 48.1 & 37.8 & 40.2 & 36.3 & 55.4 & 52.9 & 57.5 & 41.4 & 58.8 & 33.5 & 43.7\\
InternVideo2.5 & 8B & 64 & 44.2 & 43.9 & 45.2 & 39.4 & 42.4 & 37.5 & 52.7 & 49.1 & 55.5 & 42.2 & 53.0 & 37.4 & 43.8\\
Qwen3-VL-8B & 8B & 64 & 44.6 & 43.1 & 48.7 & 40.1 & 43.6 & 37.9 & 50.9 & 49.2 & 52.1 & 42.9 & 52.8 & 38.4 & 44.0\\
Doubao-Seed-1.6V & -- & 50 & 49.6 & 49.4 & 50.1 & 42.8 & 46.4 & 40.4 & 56.0 & 54.9 & 56.8 & 44.6 & 57.0 & 39.0 & 47.3\\
Qwen2.5-VL-32B & 32B & 64 & 50.3 & 49.4 & 53.0 & 40.9 & 43.1 & 39.5 & 55.4 & 50.4 & 59.3 & 46.0 & 56.4 & 41.4 & 47.3\\
Qwen3-VL-32B & 32B & 64 & 50.3 & 49.0 & 54.1 & 46.5 & 50.1 & 44.3 & 55.2 & 53.9 & 56.3 & 46.7 & 58.8 & 41.2 & 49.0\\
DirectMe (Qwen3-VL) & 8B & 1fps & 54.1 & 53.0 & 57.2 & 47.6 & 50.2 & 46.0 & 58.0 & 55.7 & 59.8 & 46.4 & 53.7 & 43.1 & 50.5\\
\midrule
\rowcolor{GroupGray}\multicolumn{16}{l}{\textit{Our complete Qwen3-VL-8B evaluation}}\\
Official base & 8B & 128 & 48.09 & 47.18 & 50.80 & 41.86 & 40.32 & 42.85 & 49.69 & 46.27 & 52.39 & 43.66 & 53.27 & 39.40 & 45.24\\
\rowcolor{MethodBlue}\textbf{\method{}} & 8B & 128 & \textbf{59.97} & \textbf{58.83} & \textbf{63.39} & \textbf{59.84} & \textbf{55.46} & \textbf{62.64} & \textbf{55.72} & \textbf{55.59} & \textbf{55.81} & \textbf{57.75} & \textbf{63.26} & \textbf{55.31} & \textbf{58.39}\\
$\Delta$ & & & +11.88 & +11.64 & +12.59 & +17.97 & +15.14 & +19.79 & +6.02 & +9.32 & +3.43 & +14.09 & +9.99 & +15.91 & +13.15\\
\bottomrule
\end{tabular}
\end{adjustbox}
\end{table*}

\subsection{Setup}
\label{sec:setup}

\stitle{Benchmarks.}
OST-Bench \citep{lin2025ost} contains 1,386 scenes and 9,528 scored online
turns of embodied indoor exploration, UCS-Bench \citep{wang2026ucs} 8,114
timestamp-causal questions over long egocentric streams, and SIS-Bench
\citep{zou2026sis} 4,856 spatial-cognition and self-awareness questions over
real UAV videos. We evaluate complete populations with greedy decoding and
each benchmark's official prompts, answer contracts, and scorers. No
benchmark test question, answer, or prediction enters training. Training
includes OST training scenes and SIS-Motion training videos; the three native
benchmarks use their held-out test populations
(Appendix~\ref{app:relational-training}). Deltas use unrounded scores.

\stitle{Training.}
We instantiate WorldIDE by fine-tuning Qwen3-VL-8B-Instruct
\citep{bai2025qwen3vl} on 45,472 source-grounded examples spanning
categorical, metric, trajectory, and observer-motion supervision.
Full-data and family-removal variants share backbone, initialization, and
update budget.
WorldIDE and the joint control share records but not family allocations.
Data composition and optimization settings appear in
Appendices~\ref{app:relational-training} and~\ref{app:evaluation}.

\stitle{Evaluation controls.}
Official-base, family-removal, and matched full-data joint controls assess
the complete relational recipe, supervision families, and family weighting.

\subsection{RQ1: Performance Across Benchmarks}
\label{sec:rq1}

\stitle{Matched gains across three spatial benchmarks.}
Table~\ref{tab:main-results} shows gains of 12.8--13.8 points over the
matched official base on every benchmark, and the scores exceed all listed
published references, which follow their own observation and inference
protocols. The gains are broad: every OST subtype and every UCS category and
subcategory improves.

\stitle{Temporal and viewpoint-dependent gains.}
On OST, the subtypes that relate the agent to objects across turns gain most:
direction estimation more than doubles (19.01 to 44.17), and order, temporal
localization, and counting rise by 17.8--19.1 points
(Table~\ref{tab:ost-main}). On UCS, position gains 17.97 points, led by
relative position (+19.79), and trajectory and category also improve
(Table~\ref{tab:ucs-main}); on SIS, self-awareness perception and memory gain
38.4 and 26.0 points (Table~\ref{tab:sis-main}). The gains thus concentrate
where answers depend on relations across time or viewpoint.
Figure~\ref{fig:case} shows one co-occurrence question asked at Turn~5:
VideoChat3, JoyAI-VL, and Mage-VL answer Turns~10, 11, and~6, none yet
observed, whereas \method{} returns the reference Turn~2 (more cases in
Appendix~\ref{app:multimethod-cases}).

\input{figures/case}

\stitle{Complementary contributions of relation families.}
Table~\ref{tab:family-ablation} shows a clear hierarchy: removing observer
motion (5.46 points on the mean) or trajectories (4.71) costs most, metric
geometry helps on every benchmark (0.50--0.75), and every removal lowers the
mean. The hierarchy mirrors where the gains concentrate: removing observer
motion costs 11.24 points on SIS and 5.72 on UCS, and removing trajectories
costs 12.77 on OST, matching the observer-relative and temporal demands of SIS
self-awareness, UCS position, and OST temporal localization. With the update
budget fixed, each drop measures a family's contribution within this recipe.
The joint control reaches \JointMean{}, within 0.09 of \method{}, so the
principal answer-level gain belongs to the full relational training recipe.

\stitle{Cross-operation distinguishability.}
On 626 fixed history pairs that agree under one operation but differ under
another, PairAcc credits a pair only if both distinguishing answers are
correct. Under identical native readout, pooled PairAcc rises from 7.83\% for
Base to 14.06\% for Joint and 13.74\% for \method{}
(Table~\ref{tab:pairacc-main}). Pooling weights all pairs equally (383 sector,
137 first-seen, and 106 threshold), so Joint's sector and threshold scores
offset \method{}'s higher first-seen PairAcc (37.96\% versus 34.31\%); the
\method{}--Joint intervals include zero in every group
(Appendix~\ref{app:pairacc-final}, which also reports single-history
accuracies and paired uncertainty).

\stitle{Relation-specific learning effects.}
The gain over Base is operation-dependent. On first-seen distinctions,
\method{} raises PairAcc from 13.87\% to 37.96\% (+24.09; 95\% visit-bootstrap
CI $[9.01,41.03]$) and single-history accuracy from 39.42\% to 63.14\%; on
signed-sector distinctions, PairAcc (3.39\% to 3.13\%) and accuracy (22.85\%
to 23.11\%) barely move, so both metrics localize the gain to temporal
readout. The trajectory control ties this asymmetry to supervision: the
\method{}-minus-no-trajectory difference is 13.14 points for first-seen and
$-1.57$ for sector, a 14.71-point interaction (95\% CI $[1.06,29.77]$). The two
readouts make different demands: first-seen locates an entity's appearance in
the observation sequence, whereas sector resolves direction in an
observer-relative frame. These exploratory contrasts thus link supervision
composition to which relations become recoverable, giving the benchmark gains
an operation-level interpretation (matching rules in
Appendix~\ref{app:pairacc-final}).

\begin{table*}[t]
\caption{\textbf{Complete SIS-Bench results} (\%). Spatial, Self, and the
self-awareness levels are unweighted task macros; Overall is
question-weighted. Published rows use their reported protocols; final block:
our matched evaluation.}
\vspace{-2mm}
\label{tab:sis-main}
\centering
\fontsize{7}{8}\selectfont
\setlength{\tabcolsep}{4.8pt}
\renewcommand{\arraystretch}{0.96}
\begin{adjustbox}{max width=\textwidth}
\begin{tabular}{lccc|rrrrrrrr|ccc}
\toprule
& \multicolumn{3}{c}{Aggregate}
& \multicolumn{8}{c}{Spatial Cognition}
& \multicolumn{3}{c}{Self-Awareness}\\
\cmidrule(lr){2-4}\cmidrule(lr){5-12}\cmidrule(lr){13-15}
Model & \shortstack{Spatial\\Avg.} & \shortstack{Self\\Avg.} & Overall
& \shortstack{Obj.\\Exist.} & \shortstack{Obj.\\Attr.}
& \shortstack{Rel.\\Dir.} & \shortstack{Land.\\Order}
& \shortstack{Land.\\Recall} & \shortstack{Pos.\\Rel.}
& \shortstack{Spat.\\Cons.} & \shortstack{ST\\Cons.}
& Perc. & Mem. & Reason.\\
\midrule
\rowcolor{GroupGray}\multicolumn{15}{l}{\textit{Representative published MLLMs}}\\
Gemini-3-Flash & 81.3 & 61.7 & 71.6 & 97.2 & 80.6 & 75.0 & 94.4 & 84.0 & 89.7 & 71.8 & 57.7 & 66.5 & 63.2 & 57.7\\
Kimi-2.5 & 81.5 & 60.8 & 71.0 & 97.0 & 78.0 & 80.5 & 98.0 & 81.9 & 87.3 & 76.4 & 53.1 & 50.9 & 65.4 & 61.2\\
Qwen3.5-Plus & 81.5 & 61.1 & 70.1 & 97.6 & 76.2 & 78.5 & 97.7 & 83.3 & 87.7 & 75.9 & 54.8 & 52.8 & 62.5 & 63.9\\
GPT-5.4 & 77.8 & 61.9 & 70.0 & 98.4 & 74.9 & 67.0 & 94.8 & 78.6 & 86.9 & 64.1 & 57.3 & 57.3 & 63.8 & 62.3\\
Qwen3-VL-8B-Instruct & 77.8 & 50.7 & 64.6 & 97.2 & 82.2 & 73.5 & 95.8 & 74.9 & 82.9 & 64.6 & 51.0 & 55.1 & 46.2 & 53.0\\
InternVL3.5-8B & 72.2 & 47.5 & 61.1 & 96.1 & 76.5 & 70.5 & 86.3 & 72.0 & 73.8 & 55.4 & 47.3 & 56.1 & 42.1 & 48.5\\
Doubao-Seed-1.8 & 79.8 & 61.0 & 70.6 & 97.4 & 72.6 & 76.5 & 96.1 & 84.4 & 87.7 & 66.2 & 57.3 & 43.7 & 72.2 & 58.5\\
Qwen3-VL-8B-Thinking & 77.2 & 52.2 & 64.7 & 95.5 & 76.0 & 72.5 & 95.1 & 78.6 & 79.8 & 64.6 & 55.6 & 53.4 & 48.2 & 55.4\\
\midrule
\rowcolor{GroupGray}\multicolumn{15}{l}{\textit{Our complete Qwen3-VL-8B evaluation}}\\
Official base & 77.6 & 51.8 & 65.1 & \textbf{97.8} & 82.4 & 76.0 & 91.2 & 74.7 & 83.3 & 62.6 & \textbf{52.7} & 53.6 & 51.3 & \textbf{51.5}\\
\rowcolor{MethodBlue}\textbf{\method{}} & \textbf{81.2} & \textbf{66.1} & \textbf{77.9} & 97.6 & \textbf{89.1} & \textbf{77.0} & \textbf{97.1} & \textbf{84.9} & \textbf{87.3} & \textbf{65.1} & 51.5 & \textbf{92.0} & \textbf{77.3} & 41.9\\
$\Delta$ & +3.6 & +14.3 & +12.8 & $-$0.2 & +6.7 & +1.0 & +5.9 & +10.2 & +4.0 & +2.5 & $-$1.2 & +38.4 & +26.0 & $-$9.6\\
\bottomrule
\end{tabular}
\end{adjustbox}
\end{table*}

\begin{table}[t]
\centering
\caption{\textbf{Contribution of relational supervision families} (\%). Each
ablation removes one family under a fixed update budget.}
\vspace{-2mm}
\label{tab:family-ablation}
\footnotesize
\setlength{\tabcolsep}{4pt}
\renewcommand{\arraystretch}{1.1}
\begin{tabular}{@{}lrrrr@{}}
\toprule
Training & OST & UCS & SIS & Mean \\
\midrule
\rowcolor{MethodBlue}
\method{} & 64.55 & 58.39 & 77.90 & \textbf{66.95} \\
w/o categorical & 66.60 & 53.91 & 77.35 & 65.95 \\
w/o metric & 63.80 & 57.70 & 77.40 & 66.30 \\
w/o trajectory & 51.78 & 58.37 & 76.57 & 62.24 \\
w/o observer motion & 65.12 & 52.67 & 66.66 & 61.49 \\
\bottomrule
\end{tabular}
\end{table}

\begin{table}[t]
\centering\footnotesize
\caption{\textbf{Cross-operation distinguishability}: native PairAcc (\%) on
all 626 fixed pairs; T/C denote trajectory/categorical removal.
Exploratory semantic groups and paired intervals: Appendix~\ref{app:pairacc-final}.}
\label{tab:pairacc-main}
\setlength{\tabcolsep}{3pt}
\begin{tabular}{@{}lrrrrrr@{}}
\toprule
Readout & $N$ & Base & Joint & WorldIDE & w/o T & w/o C \\
\midrule
First-seen & 137 & 13.87 & 34.31 & 37.96 & 24.82 & 36.50 \\
Threshold & 106 & 16.04 & 23.58 & 20.75 & 16.98 & 19.81 \\
Sector & 383 & 3.39 & 4.18 & 3.13 & 4.70 & 5.48 \\
Pooled & 626 & 7.83 & 14.06 & 13.74 & 11.18 & 14.70 \\
\bottomrule
\end{tabular}
\end{table}

\begin{table}[t]
\caption{\textbf{Native state reuse and source dependence} on complete
SIS-Bench (4,856 questions). Accuracy changes are in percentage points;
encoding reduction is relative.}
\label{tab:functional-tests}
\centering
\footnotesize
\setlength{\tabcolsep}{3pt}
\renewcommand{\arraystretch}{1.05}
\begin{tabular}{@{}lrrr@{}}
\toprule
\multicolumn{4}{@{}l}{\textit{(a) Shared-prefix execution}}\\
Metric & Direct & Shared & $\Delta$ \\
\midrule
Accuracy & \WorldIDEShared{} & \WorldIDEShared{} & 0.00 \\
Visual-prefix encodes & 4,856 & 1,601 & $-\SharedEncodeReduction\%$ \\
\midrule
\multicolumn{4}{@{}l}{\textit{(b) Source substitution}}\\
Metric & Correct & Wrong & $\Delta$ \\
\midrule
Accuracy & \WorldIDEShared{} & \WorldIDEDeranged{} & $-25.74$ \\
\bottomrule
\end{tabular}
\end{table}

\subsection{RQ2: Reusable, Source-Bound State}
\label{sec:rq2}
\stitle{One encoding serves every question about a video.}
If \method{} holds what it learns as one state per video (C3;
Figure~\ref{fig:overall}b), a single encoding should serve all of that video's
questions. On complete SIS, encoding each video once and reusing the state
leaves accuracy unchanged at 77.90\% while cutting visual-prefix encodings
from 4,856 to 1,601 (67.03\%), about three questions per encoding
(Table~\ref{tab:functional-tests}a). The exact agreement shows that the state
carries nothing question-specific, so it can be intervened on directly.

\stitle{Answers are bound to the observed video.}
An answer-blind cyclic permutation then gives each question another video's
state, with the question, checkpoint, decoder, and scorer fixed. Accuracy
falls from 77.90\% to 52.16\%: the 25.74-point loss removes a third of
\method{}'s accuracy (Table~\ref{tab:functional-tests}b), and \method{} now
trails the base model reading its own video (65.09\%) by 12.93 points,
suggesting that its gain flows through the observed history rather than
through question or option priors. Two additional training variants replicate
both results, with reuse accuracy unchanged to within 0.02 points and drops of
24.22 and 23.58 points (Appendix~\ref{app:shared-state}). In short, \method{}
keeps one reusable state per video, and its answers depend on which video that
state encodes.

\subsection{RQ3: Explicit Geometric Readout}
\label{sec:rq3}

\stitle{Geometric properties at similar answer accuracy.}
Joint and WorldIDE have nearly identical benchmark means (66.86 and 66.95),
yet their representations differ under matched probing. On 14,435 content
queries at 514 endpoints, WorldIDE lowers disagreement between predicted
distance and displacement magnitude from 0.605 to 0.546\,m
(Table~\ref{tab:state-recoverability}). The reduction is 0.060\,m using
unrounded values, with paired 95\% visit-bootstrap CI $[0.031,0.088]$.
This residual measures internal field agreement. Source-distance MAE is
0.800\,m for Joint and 0.785\,m for WorldIDE; the paired difference
interval includes zero.
Appendix~\ref{app:state-joint-final} reports both measures jointly.

The comparison holds relational examples fixed and changes their relative
training emphasis. It therefore examines how family allocation relates to
geometric agreement after learning. Because probe capacity, fitting protocol,
and evaluation population are matched, the residual supplies a common
reference for comparing recovered fields. Its value lies in distinguishing
geometric properties that remain hidden when two models are summarized by
nearly identical benchmark means.

The 130 evaluation captures are held out from probe-head fitting but occur
among encoder SFT sources. Strict joint recovery remains limited
(Table~\ref{tab:state-recoverability}); additional probe and cross-operation
diagnostics appear in Appendices~\ref{app:state-plane} and
\ref{app:diagnostic-boundaries}.

\stitle{Discussion.}
PairAcc tests distinctions across histories; matched geometric probes test
compatibility within a history.

\stitle{Why geometric agreement matters.}
Question-level accuracy treats each requested quantity separately, although
distance and displacement describe the same spatial configuration. Individually plausible
predictions can violate the geometric constraint linking them. The State
Plane tests this constraint through independently recovered quantities at
the same endpoint. Geometric compatibility therefore complements answer
correctness by measuring agreement among spatial readouts tied to a common
configuration.

Distance and displacement are predicted independently; the executor does not
enforce their agreement. Matched probing therefore tests whether separately
recovered fields respect a shared geometric constraint, complementing the
benchmarks' assessment of answer quality.

Cross-history discrimination and within-history compatibility need not
improve together, and compatible fields may disagree with source geometry.
Our findings show operation-dependent history distinctions and distance-field
agreement differences at similar benchmark accuracy. Relational state
identification links these properties to operations specifying what to
preserve across histories and recover within them.

\begin{table}[t]
\centering
\caption{\textbf{Task-blind spatial-state recoverability}: 14,435 queries,
514 endpoints. Accuracies are \%; entity/endpoint joint requires all associated
operations. Residuals use meters and degrees. Protocol:
Appendix~\ref{app:taskblind-plane}; visit-bootstrap comparisons:
Appendix~\ref{app:state-joint-final}.}
\label{tab:state-recoverability}
\footnotesize
\setlength{\tabcolsep}{1.6pt}
\renewcommand{\arraystretch}{1.1}
\begin{tabular}{@{}lrrrrr@{}}
\toprule
Encoder & Query & Entity joint & Endpoint & $e_d\downarrow$ & $e_\theta\downarrow$ \\
\midrule
Official base & 52.64 & 1.80 & 0.19 & 0.621 & 15.06 \\
Joint & 53.15 & \textbf{2.35} & 0.00 & 0.605 & 16.46 \\
\rowcolor{MethodBlue}
\textbf{\method{}} & \textbf{53.70} & \textbf{2.35} & \textbf{0.39} &
\textbf{0.546} & \textbf{14.97} \\
\bottomrule
\end{tabular}
\end{table}

%% file: figures/case.tex
\begin{figure}[!t]
\centering
\input{figures/case-ost/case-tikz.tex}
\captionsetup{skip=0pt}
\vspace{-2mm}
\caption{\textbf{Three external models name turns not yet observed;
\method{} names Turn~2.} The Figure~\ref{fig:teaser} episode at Turn~5; the
sink and plant co-occur in Turn~2 (top). Only observations through Turn~5
are available to this query. More in
Appendix~\ref{app:multimethod-cases}.}
\vspace{-2mm}
\label{fig:case}
\end{figure}
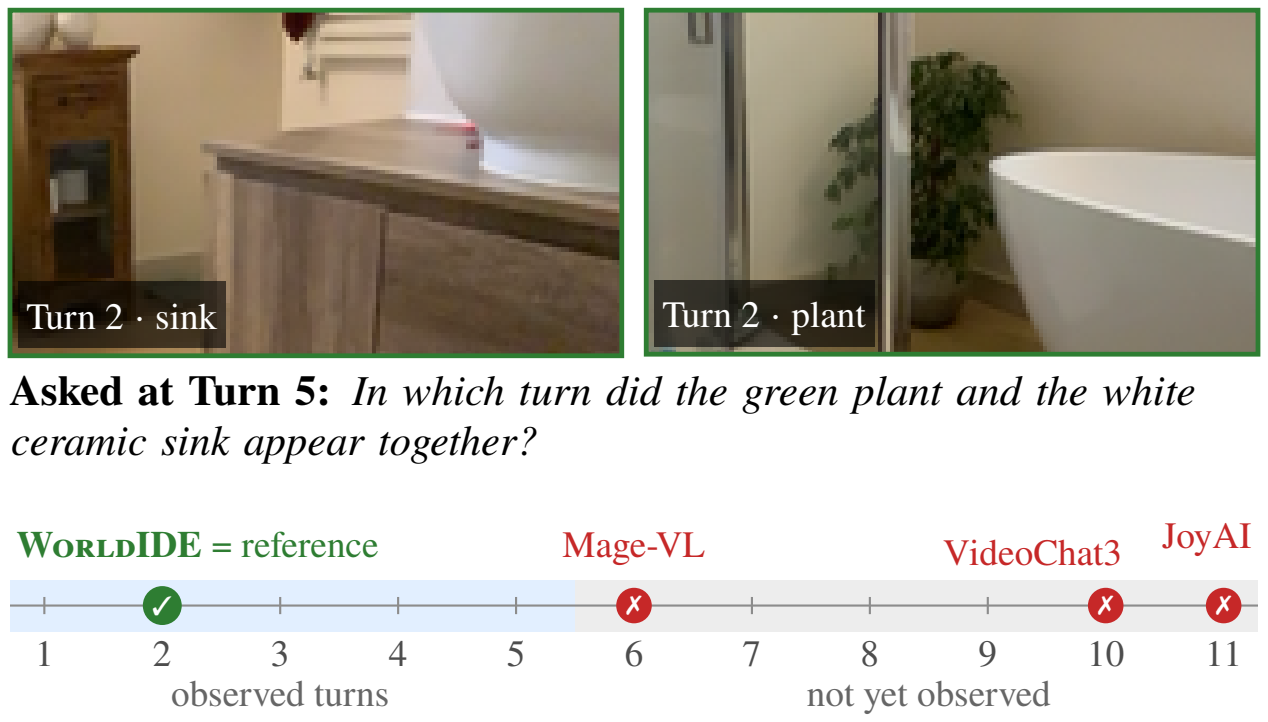

%% file: figures/case-ost/case-tikz.tex
\begin{tikzpicture}[x=\columnwidth/218, y=\columnwidth/218,
    trim left=0pt, trim right=\columnwidth,
    lab/.style={font=\scriptsize, inner sep=1pt},
    tag/.style={font=\scriptsize, text=white, fill=black, fill opacity=0.6,
                text opacity=1, inner sep=1.6pt, anchor=south west,
                text height=1.6ex, text depth=0.45ex},
    ok/.style={circle, draw=CaseGreen, fill=CaseGreen, text=white,
               inner sep=0.6pt, font=\tiny},
    bad/.style={circle, draw=CaseRed, fill=CaseRed, text=white,
                inner sep=0.6pt, font=\tiny}]
  \node[anchor=north west, inner sep=0] (sink) at (0,0)
    {\includegraphics[width=0.4885\columnwidth]{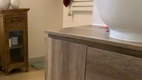}};
  \node[anchor=north east, inner sep=0] (plant) at (218,0)
    {\includegraphics[width=0.4885\columnwidth]{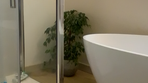}};
  \draw[CaseGreen, line width=0.9pt] (sink.south west) rectangle (sink.north east);
  \draw[CaseGreen, line width=0.9pt] (plant.south west) rectangle (plant.north east);
  \node[tag] at ([shift={(1.2,1.2)}]sink.south west) {Turn 2 $\cdot$ sink};
  \node[tag] at ([shift={(1.2,1.2)}]plant.south west) {Turn 2 $\cdot$ plant};
  \node[anchor=north west, inner sep=0, text width=\columnwidth,
        font=\footnotesize, align=left] (q) at ([yshift=-3.5pt]sink.south west)
    {\textbf{Asked at Turn 5:} \emph{In which turn did the green plant and the
     white ceramic sink appear together?}};
  \def\ty{-104}
  \pgfmathsetmacro{\xa}{6}\pgfmathsetmacro{\dx}{20.6}
  \fill[MethodBlue] (0,\ty-4.5) rectangle (\xa+4.5*\dx,\ty+4.5);
  \fill[black!7]    (\xa+4.5*\dx,\ty-4.5) rectangle (218,\ty+4.5);
  \draw[black!45, line width=0.4pt] (0,\ty) -- (218,\ty);
  \foreach \k in {1,...,11} {
    \draw[black!45, line width=0.4pt] (\xa+\k*\dx-\dx,\ty-1.5) -- ++(0,3);
    \node[lab, text=black!70, anchor=north] at (\xa+\k*\dx-\dx,\ty-5) {\k};
  }
  \node[lab, text=black!60, anchor=north] at (\xa+2*\dx,\ty-12) {observed turns};
  \node[lab, text=black!60, anchor=north] at (\xa+7.5*\dx,\ty-12) {not yet observed};
  \node[ok]  (wi) at (\xa+1*\dx,\ty) {\ding{51}};
  \node[bad] (mv) at (\xa+5*\dx,\ty) {\ding{55}};
  \node[bad] (vc) at (\xa+9*\dx,\ty) {\ding{55}};
  \node[bad] (jy) at (\xa+10*\dx,\ty) {\ding{55}};
  \node[lab, anchor=south west, text=CaseGreen] at ([yshift=3pt]0,\ty+4.5)
    {\textbf{\method{}} = reference};
  \node[lab, anchor=south, text=CaseRed] at ([yshift=3pt]mv.north) {Mage-VL};
  \node[lab, anchor=south east, text=CaseRed] at ([shift={(4,3pt)}]vc.north) {VideoChat3};
  \node[lab, anchor=south east, text=CaseRed] at (218,\ty+7.5) {JoyAI};
\end{tikzpicture}

%% file: sections/related_work.tex
\section{Related Work}

\stitle{Streaming video memory.}
Long-video methods retain evidence through token memories
\citep{song2024moviechat,zhang2025flashvstream}, online memory banks
\citep{he2024malmm}, and adaptive spatiotemporal compression
\citep{shen2025longvu}. Streaming systems support later queries \citep{huang2025videochatonline,zeng2026onestreamer}; DirectMe
maintains user-centric spatial memory \citep{wang2026ucs}.
ViSS studies visible-state supervision and causal view selection as separate
factors in streaming understanding \citep{hu2026viss}.

\stitle{Spatial grounding.}
3D question answering and grounding connect geometry with language
\citep{azuma2022scanqa,wang2024embodiedscan,hong2023threedllm,huang2024leo}.
Spatial vision-language models learn from depth, 3D geometry, and relational
targets
\citep{chen2024spatialvlm,cheng2024spatialrgpt,cai2025spatialbot,mao2025spatiallm,ray2025sat,wu2025spatialmllm}.
Reconstruction and scene memory
\citep{wang2025vggt,Gu_2024,Yang_2025_CVPR_3DMem} support structured
reasoning \citep{ma2025spatialreasoner}; spatial benchmarks measure
answer-level performance
\citep{kamath-etal-2023-whats,Fu_2024,du-etal-2024-embspatial,Yang_2025_CVPR}.
Cross-View Relations supervises correspondence, verification, and localization
across views \citep{jeong2026xvr}. ChronoReplay balances heterogeneous spatial
targets through chronological episodes \citep{hu2026chronoreplay}. Our analysis additionally asks which
cross-history distinctions remain behaviorally accessible and which
geometric relationships can be recovered under matched probing.

\stitle{Functional state and world models.}
Predictive state representations, DeepMDP, and bisimulation characterize
state through observable or behavioral distinctions
\citep{singh2004psr,gelada2019deepmdp,zhang2021bisimulation}.
Probe controls help distinguish recoverable information from probe learning \citep{hewitt-liang-2019-designing,belinkov-2022-probing};
amnesic interventions examine its use in downstream behavior
\citep{elazar2021amnesic}.
World models learn predictive dynamics
\citep{hafner2025dreamerv3,bruce2024genie,assran2025vjepa2}
or executable programs
\citep{tang2024worldcoder,chen2026codeworldmodel,wang2026codeasworlds};
programmatic vision composes grounded operations
\citep{gupta2023visprog,suris2023vipergpt}.
Additional related work appears in Appendix~\ref{app:related}.

%% file: sections/conclusion.tex
\section{Conclusion}

Spatial understanding requires more than correct individual answers:
histories may agree on one spatial relation yet differ in temporal
order, viewpoint, or other properties that later reasoning needs. \method{}
supervises complementary, source-verifiable relations to preserve these
distinctions.
Across three benchmarks, relational supervision improves mean accuracy by
13.25 points, with operation-dependent gains in history distinguishability;
probes show that models with similar accuracy can differ in the geometric
compatibility of recovered quantities. Spatial representations should thus be
judged not only by isolated answers but by whether they distinguish histories
with different relational consequences and recover compatible geometry within
an observation.

%% file: sections/limitations.tex
\section*{Limitations}

Two boundaries are visible in our results. First, relational supervision
improves observed spatial reasoning while future-action reasoning remains limited: SIS path
planning falls from 45.6 to 27.9, and the same drop appears in the joint
control and two other trained variants
(Appendix~\ref{app:diagnostic-boundaries}). The
principle suggests a possible extension. Planning questions are operations the
state was never asked to support; adding a family whose targets are executed
over candidate futures from the same source geometry, such as reachability
or next-waypoint relations, would add them to the target signature; their effect would require evaluation.

Second, answer accuracy and explicit recoverability use different protocols.
The State Plane predicts distance, displacement, and angle independently; \method{} lowers the distance residual
(Table~\ref{tab:state-recoverability}) but residuals remain nonzero, and
endpoints whose every query is correct remain rare. The head was kept separate and the backbone
frozen to measure recoverability under matched probe training. The task-blind
validation captures are held out from head fitting but occur in encoder SFT
sources; they do not establish unseen-source generalization. Future studies can
supervise typed fields alongside the relational objective, add cross-field
consistency constraints (distance and displacement agreeing,
bearing equivariant under the declared frame transforms).

\section*{Ethics Statement}

Persistent spatial memory may retain sensitive information about people and environments. A deployment should minimize stored observations, restrict state fields to the task, retain provenance and deletion controls, and avoid using inferred identity as an entity key. Our operator schema does not execute arbitrary model-generated code. The reported training sources and benchmark protocols are used for research evaluation; no held-out benchmark answers or predictions are inserted into training data.

%% file: sections/appendix.tex
\suppressfloats[t]

\section{Functional Identifiability of Spatial State}
\label{app:theory}

Relational state identification characterizes a spatial representation through
the operations it can support. We formalize the information required for
shared-state readout and clarify the relationship between operation-conditioned
supervision and the functional properties evaluated in WorldIDE. The analysis
concerns exact sufficiency on a declared operation family; its realization in
learned representations is assessed empirically.

\subsection{States Defined by Observable Operations}

Let $\mathcal H$ be the space of causal visual histories and $\mathcal R$ a
family of grounded spatial and temporal operations. Each $r\in\mathcal R$
specifies its entity arguments, reference frame, temporal scope, and output
type. For a history $h$, let $y_r(h)$ denote its source-verifiable target.
Operations outside the available evidence may return a typed unknown value.

The relation signature of a history is
\begin{equation}
  \rho_{\mathcal R}(h)
  =
  \bigl(y_r(h)\bigr)_{r\in\mathcal R}.
  \label{eq:app-relation-signature}
\end{equation}
It induces an equivalence relation
\begin{equation}
  h\sim_{\mathcal R}h'
  \quad\Longleftrightarrow\quad
  \rho_{\mathcal R}(h)=\rho_{\mathcal R}(h').
  \label{eq:app-state-equivalence}
\end{equation}
Two histories are equivalent when every declared operation produces the same
result. Their physical configurations may differ in properties outside
$\mathcal R$. Equal distances, for example, permit opposite bearings under a
distance-only signature, whereas adding signed bearing distinguishes the
states.

The ideal operation-relative state is the equivalence class
\begin{equation}
  S_{\mathcal R}^{\star}(h)
  =
  [h]_{\sim_{\mathcal R}}.
  \label{eq:app-functional-state}
\end{equation}
This state preserves exactly the distinctions observable through
$\mathcal R$. Expanding the operation family refines its equivalence classes:
\begin{equation}
  \mathcal R_1\subseteq\mathcal R_2
  \quad\Longrightarrow\quad
  [h]_{\sim_{\mathcal R_2}}
  \subseteq
  [h]_{\sim_{\mathcal R_1}}.
  \label{eq:app-refinement}
\end{equation}
Relational identification therefore specifies state information at the
resolution of a declared operation family while leaving the latent
parameterization unconstrained.

\subsection{When Can One Representation Support All Readouts?}

Consider a query-independent encoder $E:\mathcal H\rightarrow\mathcal Z$ and a
family of readers $\{D_r\}_{r\in\mathcal R}$. We call $E$
\emph{$\mathcal R$-sufficient} when every operation can recover its target from
the same representation:
\begin{equation}
  D_r(E(h))=y_r(h),
  \quad
  \forall h\in\mathcal H,\;
  \forall r\in\mathcal R.
  \label{eq:app-sufficiency}
\end{equation}

\stitle{Proposition 1 (Operational sufficiency).}
For a deterministic encoder $E$, such a reader family exists if and only if
\begin{equation}
  E(h)=E(h')
  \quad\Longrightarrow\quad
  h\sim_{\mathcal R}h'.
  \label{eq:app-sufficiency-condition}
\end{equation}

\stitle{Proof.}
If the readers exist, identical encodings produce identical outputs for every
operation, establishing the forward implication. Conversely, suppose
Equation~\ref{eq:app-sufficiency-condition} holds. For each $z$ in the image of
$E$, choose a history $h_z$ with $E(h_z)=z$ and define
$D_r(z)=y_r(h_z)$. All histories mapped to $z$ have the same relation
signature, so $D_r$ is well-defined and recovers every target.
\hfill$\square$

The proposition gives a precise criterion for reusable state: the encoder must
distinguish histories whenever the declared operations require different
results. Consider two histories satisfying
\begin{equation}
  y_{r_a}(h)=y_{r_a}(h'),
  \qquad
  y_{r_b}(h)\neq y_{r_b}(h').
  \label{eq:app-cross-operation-ambiguity}
\end{equation}
Supervision through $r_a$ permits $E(h)=E(h')$. If this collision occurs, the
representation is insufficient for $r_b$. Under a balanced exact-match
evaluation of these histories, any deterministic reader of the shared encoding
achieves at most $50\%$ accuracy on $r_b$.

We refer to this potential representational failure as \emph{query-induced
state aliasing}. The distinction is between histories that agree on one
supervised answer and histories that are equivalent under the complete
operation family. Complementary relational targets refine the distinctions
that supervision encourages the encoder to retain.

\subsection{Learning and Functional Evaluation}

WorldIDE optimizes source-grounded, operation-conditioned predictions
(Appendix~\ref{app:relational-training}):
\begin{equation}
  \mathcal L_{\mathrm{rel}}(\theta)
  =
  \mathbb E_{(h,r,y)}
  \left[-\log p_\theta(y\mid h,r)\right].
  \label{eq:app-relational-objective}
\end{equation}
Proposition~1 characterizes the information required for exact operational
sufficiency. The training objective encourages these distinctions without
directly enforcing an explicit state factorization.

We evaluate learned representations through two complementary interfaces.
Native shared-prefix readout examines repeated access and dependence on the
observation history. An auxiliary State Plane measures the recoverability of
typed spatial quantities through deterministic operations. Together, these
interfaces connect the functional state criterion to measurable properties
of learned spatial representations.

\section{Controlled Comparisons and Attribution}
\label{app:attribution}

Table~\ref{tab:attribution-matrix} maps each controlled comparison to its
measured property and experimental protocol.

\begin{table*}[t]
\centering
\caption{\textbf{Controlled comparisons and evidence attribution.} Each
contrast retains its declared checkpoint and evaluation protocol. Accuracy
changes are in percentage points (pp). Auxiliary State Plane heads are fitted
separately under matched settings.}
\label{tab:attribution-matrix}
\footnotesize
\setlength{\tabcolsep}{4pt}
\renewcommand{\arraystretch}{1.15}
\begin{tabularx}{\textwidth}{
@{}
>{\raggedright\arraybackslash}p{0.19\textwidth}
>{\raggedright\arraybackslash}p{0.28\textwidth}
>{\raggedright\arraybackslash}p{0.18\textwidth}
>{\raggedright\arraybackslash}X
@{}
}
\toprule
Contrast & Controlled factors & Result & Supported conclusion \\
\midrule
\rowcolor{MethodBlue}
\multicolumn{4}{@{}l}{\textbf{(a) Relational learning and recoverability}}\\
Base $\rightarrow$ Joint
& Benchmark evidence, prompts, decoding, scoring
& $53.70\rightarrow66.86$
& Aggregate gain from the relational training recipe \\
Joint $\rightarrow$ WorldIDE
& 45,472 rows, initialization, optimizer, seed, update budget
& $66.86\rightarrow66.95$
& Increment associated with family reweighting \\
Base $\rightarrow$ WorldIDE Plane
& Probe data, architecture, optimization; separately fitted heads
& $61.65\rightarrow63.57$
& Spatial recoverability under matched probing \\
\midrule
\rowcolor{MethodBlue}
\multicolumn{4}{@{}l}{\textbf{(b) Reusable visual state}}\\
Direct $\rightarrow$ Shared
& SIS checkpoint, observations, questions, decoder, scorer
& $77.90\rightarrow77.90$; $-67.03\%$ encodes
& Accuracy-preserving reuse of visual-prefix computation \\
Correct $\rightarrow$ Wrong source
& Checkpoint, target question, decoding, scorer
& $77.90\rightarrow52.16$
& Source-dependent native readout \\
\midrule
\rowcolor{MethodBlue}
\multicolumn{4}{@{}l}{\textbf{(c) Explicit state and component diagnostics}}\\
Correct $\rightarrow$ Wrong Plane
& Frozen encoder and head, operation, arguments, target
& $63.57\rightarrow48.49$
& Source-sensitive typed spatial recovery \\
Reference-frame rotation
& Predicted state fields, deterministic executor
& $1,696/1,696$ valid
& Correct execution of declared geometric laws \\
Cross-operator diagnostic
& Source-disjoint screen, fixed evaluation protocol
& Compatibility $-1.56$; direct $+1.57$
& Joint improvement criterion remains unmet \\
\bottomrule
\end{tabularx}
\end{table*}

\paragraph{Attribution of empirical evidence.}
The comparisons distinguish the complete relational training recipe from
family weighting. Base-to-Joint measures the combined contribution of
additional relational supervision, whereas Joint-to-WorldIDE isolates the
normalized family allocation under matched training conditions.

The remaining contrasts evaluate complementary representation properties.
Shared-prefix inference measures repeated access, source substitution measures
dependence on observation history, and matched auxiliary probes examine
explicit spatial recoverability. Detailed results and controls appear in
Appendices~\ref{app:relational-training}, \ref{app:shared-state}, and
\ref{app:state-plane}.

\section{Source-Grounded Relational Supervision}
\label{app:relational-training}

WorldIDE learns spatial distinctions from relational examples tied to source
observations and annotations. This appendix specifies target provenance,
operation construction, and the training mixture used in
Section~\ref{sec:relational-learning}.

\subsection{Source-Grounded Operation Construction}

Let $h=o_{\leq t}$ denote a causal visual history and $g_h$ the corresponding
source annotations available to the data-generation procedure. For programmatically evaluated operations, each instance associates the
history with an operation $r$ and its geometric target:
\begin{equation}
  y_r(h)=\operatorname{Eval}(r,g_h).
  \label{eq:app-source-target}
\end{equation}
An operation specifies entity arguments, reference frame, temporal scope,
and output contract. The student receives the visual history and operation
request; geometric annotations supply the computed target.

\stitle{Target provenance.}
Computed targets come from source geometry; source-annotated targets retain the released labels. We keep this distinction explicit because annotation checks do not independently establish geometric correctness.

\stitle{Relational coverage.}
We construct four complementary families of operation targets.

\emph{Categorical relations} supervise qualitative spatial configurations,
including relative arrangements and entity relations. These targets preserve
distinctions expressed through discrete spatial decisions.

\emph{Metric geometry} supervises quantitative spatial properties, including
displacement magnitudes and signed directions. These operations distinguish
configurations that may share a categorical relation while differing in
metric structure.

\emph{Object trajectories} supervise changes in spatial state across
observations. Their temporal arguments bind each target to the corresponding
causal history and observation interval.

\emph{Observer motion} supervises motion of the observing camera and the
reference-frame changes that accompany it. These targets connect spatial
observations acquired from different viewpoints.

\stitle{Training sources and visual inputs.}
Categorical supervision uses synthetic spatial QA; metric targets use ScanNet and ARKit geometry. Trajectory targets use OST training scenes, and observer-motion targets use released SIS-Motion annotations. These sources provide complementary spatial and temporal relations over causal visual inputs.

\stitle{Causal grounding.}
Each target is associated with an explicit observation endpoint. The student
input contains the causal visual evidence available at that endpoint. Entity
identities, reference frames, and temporal arguments are resolved against the
corresponding source annotations during target construction.

For computed geometric targets, the generation procedure evaluates source
geometry while keeping its annotated values outside the student input.
Released QA targets retain the source annotations described above.

\stitle{Causal-history coverage.}
Only 2.61\% of normalized visual prefixes have multiple textual requests. Complementary constraints are therefore learned primarily across the training distribution, rather than from all four families at every observation endpoint.

\subsection{Relational Training Objective}

The resulting training population contains 45,472 records.
Table~\ref{tab:relational-mixture} reports its composition and the normalized
family contributions used by the WorldIDE checkpoint.

\begin{table}[t]
\centering
\caption{\textbf{Source-grounded relational training.} Record counts describe
the training population; loss masses specify the normalized contribution of
each supervision family to the training objective.}
\label{tab:relational-mixture}
\small
\setlength{\tabcolsep}{4pt}
\renewcommand{\arraystretch}{1.12}
\begin{tabular}{@{}lrr@{}}
\toprule
Supervision family & Records & Loss mass \\
\midrule
Categorical relations & 30,635 & $1/12$ \\
Metric geometry & 2,545 & $1/3$ \\
Object trajectories & 6,381 & $1/3$ \\
Observer motion & 5,911 & $1/4$ \\
\midrule
\textbf{Total} & \textbf{45,472} & $\mathbf{1}$ \\
\bottomrule
\end{tabular}
\end{table}

\stitle{Operation-conditioned learning.}
The visual-language model predicts a source-derived target conditioned on its
causal history and grounded operation:
\begin{equation}
  \mathcal L_{\mathrm{rel}}(\theta)
  =
  \mathbb E_{(h,r,y)\sim\mathcal D_{\mathrm{rel}}}
  \left[-\log p_\theta(y\mid h,r)\right].
  \label{eq:app-relational-sft}
\end{equation}

The operation families differ substantially in record count and target length.
We implement the effective training objective through normalized family losses:
\begin{equation}
\begin{aligned}
  \mathcal L(\theta)
  &=
  \sum_{f\in\mathcal F}
  \alpha_f\overline{\ell}_f(\theta),\\
  \overline{\ell}_f(\theta)
  &=
  \frac{
    \sum_{x\in\mathcal D_f}w_x\ell(\theta;x)
  }{
    \sum_{x\in\mathcal D_f}w_x
  },
\end{aligned}
\label{eq:app-family-objective}
\end{equation}
where $\mathcal D_f$ is the set of records in family $f$, $w_x$ denotes its
within-family weight, and $\ell(\theta;x)$ is the sample-normalized
assistant-token prediction loss. The family coefficients satisfy
$\sum_{f\in\mathcal F}\alpha_f=1$.

The family coefficients are
\begin{equation}
  \boldsymbol{\alpha}
  =
  \left(
  \frac{1}{12},
  \frac{1}{3},
  \frac{1}{3},
  \frac{1}{4}
  \right),
  \label{eq:app-family-weights}
\end{equation}
ordered as categorical, metric, trajectory, and observer-motion supervision.

These coefficients allocate training emphasis independently of raw family
size. In particular, the largest categorical family receives a smaller
fraction of normalized loss mass, while metric and trajectory operations
receive greater relative emphasis.

The training objective acts on operation-conditioned predictions from the
shared visual-language model. Its relationship to query-independent state
sufficiency is assessed through the native and explicit readout protocols, as
formalized in Appendix~\ref{app:theory}.

\subsection{Matched Training and Data Integrity}

\stitle{Matched family-weighting control.}
We distinguish the complete relational training recipe from the effect of
changing family loss mass.

The joint control uses normalized family coefficients
$(13/60,13/60,13/60,7/20)$ for categorical, metric, trajectory, and
observer-motion supervision, respectively. The first three families each
receive one third of the original pool's loss mass; their combined mass
is rescaled to 65\%, with 35\% assigned to observer motion.
WorldIDE uses the four-family allocation in
Table~\ref{tab:relational-mixture}.

Both checkpoints train on the same 45,472 records. Messages, images,
source-grounded labels, supervised assistant-token masks, within-family
weights, initialization, optimizer, training seed, and update budget are held
fixed. Only the normalized family allocation changes.

The complete three-benchmark mean increases from 66.86 to 66.95 under this
intervention. Accordingly, the matched comparison measures the incremental
effect of reweighting, while the larger improvement over the official base
reflects the complete source-grounded relational training recipe.
Appendix~\ref{app:attribution} reports these contrasts alongside their
attribution boundaries.

\stitle{Source and benchmark separation.}
Training records combine computed geometric targets and released source
annotations with their visual histories. Official held-out benchmark questions, reference
answers, task categories, model predictions, and measured evaluation scores
are excluded from training records and target-generation inputs.

Training and evaluation videos are checked for literal source overlap; the
recorded SIS training/test video overlap is zero. Source identifiers and
observation endpoints are retained to audit the correspondence between visual
evidence and relational targets.

Appendix~\ref{app:evaluation} summarizes matched training and benchmark evaluation.

\section{Shared Visual-State Execution}
\label{app:state-readout}
\label{app:shared-state}

The native path separates observation processing from question realization:
$h\mapsto Z_h\mapsto\widehat y_j$. Its state and reader remain distinct from
the auxiliary State Plane in Appendix~\ref{app:state-plane}.

\subsection{Prefix Boundary and Reuse}

The common prefix contains causal visual evidence and query-independent
context. The boundary is immediately after the final visual token and before
every question-dependent token. One prefill materializes cached key--value
representations bound to checkpoint, source video, causal boundary, and
observation policy. Each later question clones this prefix and appends its own
text suffix; questions, options, and generated answers are excluded from the
shared state.

Complete SIS evaluation contains 4,856 questions over 1,601 video prefixes.
Direct and shared arms use identical evidence, prompts, answer contracts,
greedy decoding, and official scoring. WorldIDE obtains 77.90 in both arms,
while prefix encodes decrease from 4,856 to 1,601 (67.03\%). MC-supervision
remains at 74.88; visual-supervision changes from 75.31 to 75.29. The reduction
counts prefix executions, rather than wall-clock latency.

\subsection{Source Derangement}

A fixed cyclic permutation replaces each target prefix with another video's
state. Selection is answer-blind; target question, checkpoint, decoder, parser,
and scorer remain fixed. WorldIDE falls from 77.90 to 52.16, a 25.74-point
loss. MC- and visual-supervision checkpoints fall by 24.22 and 23.58 points.
This establishes native source dependence, not explicit metric reconstruction.

\section{Executable State Plane}
\label{app:state-plane}

The auxiliary path is $h\mapsto H_h\mapsto\widehat S_h\mapsto\widetilde y_r$.
It extracts typed quantities from frozen endpoint features and evaluates them
with deterministic operators. It is trained separately from the native reader.

\subsection{Earlier Single-Track Protocol}

We extract layers $-1$, $-5$, and $-9$. Each contributes the last-token
feature, image-token mean, and attention-mask-weighted prefix mean. Their
concatenation forms $H_h$. Questions, options, and reference answers are
excluded, so all endpoint queries share one input representation.

The head applies layer normalization and two 512-dimensional GELU layers, with
dropout $0.1$ after the first. Branches predict right displacement, forward
displacement, distance, sine and cosine of bearing, 18 entity categories, and
three four-way event-turn labels. Its loss is
\begin{equation}
\begin{aligned}
\mathcal L_{\mathrm{plane}}
&=\mathcal L_{\mathrm{geom}}+0.25\mathcal L_{\mathrm{entity}}\\
&\quad+0.2\sum_{k\in\mathcal A}\mathcal L_{\mathrm{event},k}.
\end{aligned}
\label{eq:appendix-plane-loss}
\end{equation}
Continuous targets use standardized Smooth L1 loss. Each frozen encoder
receives an independently fitted head under matched training settings.

The object-track split has 3,392 training and 848 validation endpoints.
The split shares 101 physical videos and uses projected unsigned bearing
targets, so it measures single-track readout under that protocol. Every target comes from source geometry or endpoint annotations.
The fitted heads use identical data and architecture.

\subsection{Field-Level Execution}

\begin{center}
\centering
\scriptsize
\setlength{\tabcolsep}{2.5pt}
\begin{tabular}{lll}
\toprule
Query & State input & Execution \\
\midrule
Distance & $\widehat d$ & return $\widehat d$ \\
Bearing & $(\widehat s,\widehat c)$ & $\operatorname{atan2}(\widehat s,\widehat c)$ \\
Sector & bearing & four angle bins \\
Threshold & $\widehat d,\tau$ & $\widehat d\leq\tau$ \\
Entity & category logits & argmax \\
Current turn & metadata & return turn \\
Event turn & event logits & argmax \\
\bottomrule
\end{tabular}
\captionof{table}{Actual field contracts of executable State Plane queries.}
\label{tab:state-plane-operators}
\end{center}

Distance and threshold read the independent distance prediction. Bearing uses
predicted sine and cosine; sectors partition it at $-135^\circ,-45^\circ,
45^\circ,$ and $135^\circ$. Thresholds are selected from
$\{0.5,1,2,3,5\}$ meters by a fixed query hash. Event queries use event heads;
current turn reads immutable metadata. Right and forward displacement
participate in training but are not used to recompute distance or bearing.
Execution is deterministic given selected fields; consistency among independently
predicted fields is a stronger learned-state property.

The runtime validates entities, frames, units, scope, and output type. On 848
validation endpoints, 5,512 queries compare correct-source states against a
turn-matched wrong physical source while preserving head, operator, arguments,
and target.

\begin{center}
\centering
\small
\setlength{\tabcolsep}{4pt}
\begin{tabular}{lrr}
\toprule
Frozen encoder & Correct & Wrong source \\
\midrule
Official base & 61.65 & 48.93 \\
Joint checkpoint & 61.74 & 48.73 \\
WorldIDE & \textbf{63.57} & 48.49 \\
\bottomrule
\end{tabular}
\captionof{table}{Earlier single-track executable readout (\%; 848 endpoints,
5,512 queries). Each encoder uses a separately fitted matched head. This
single-track protocol differs from the task-blind multi-entity protocol in
Table~\ref{tab:state-recoverability}.}
\label{tab:appendix-plane-results}
\end{center}

Under this earlier protocol, WorldIDE improves correct-source accuracy by
1.92 points over Base, below the evaluation criterion of a two-point gain.
Source substitution lowers accuracy by 15.08 points for WorldIDE and 12.72
points for Base; WorldIDE exceeds the criterion's ten-point source-gap
requirement. The combined criterion is therefore unmet.

Reference-frame checks apply $\pm90^\circ$ observer-heading transforms to all
848 predicted validation states. Across 1,696 interventions, bearing follows
the prescribed transform, distance and current turn remain invariant, and the
inverse transform recovers the original output. These checks verify the
executor's declared reference-frame transformations.

\subsection{Task-Blind Multi-Entity Protocol}
\label{app:taskblind-plane}

Table~\ref{tab:state-recoverability} uses 14,435 content queries over 514
endpoints, 1,445 entities, and 130 ARKit captures. These comprise seven
single-entity operations with 1,445 queries each, plus inter-entity distance,
left-of, and nearer-to-camera operations with 1,440 queries each.
Encoder inputs contain a
causal image packet with no question, entity label, or operation. The head
receives the shared pooled features and a 64-dimensional entity embedding,
then applies layer normalization, 768- and 512-dimensional GELU layers, and
0.1 dropout. Outputs are signed right/forward displacement, distance,
sine/cosine of bearing, visibility, and first-/last-seen turns. Each frozen
encoder receives a separately fitted head under matched training settings.
Ambiguous same-label entities are excluded.
Visibility and event targets use the source frustum-box visibility proxy.

The 559/130 train/validation capture split is disjoint for head fitting.
All 130 validation captures occur in the main encoder's OST SFT sources,
and capture identity does not establish visit- or room-level separation.
These tests therefore measure source-conditioned accessibility.

Entity joint groups all seven operations by endpoint and entity, irrespective
of a threshold argument: the corrected success counts are 26/1,445 for Base
and 34/1,445 for both Joint and WorldIDE. Endpoint joint counts are 1/514,
0/514, and 2/514. Earlier grouping by complete query arguments split threshold
operations from the other fields and did not measure entity conjunction.
For fields sharing entity, endpoint, units, and reference frame, residuals are
\begin{equation}
 e_d=\left|\widehat d-\sqrt{\widehat{\Delta x}^{2}
                         +\widehat{\Delta z}^{2}}\right|,
 \label{eq:taskblind-distance-residual}
\end{equation}
\begin{equation}
 e_\theta=\left|\operatorname{wrap}\left(\widehat\beta-
 \operatorname{atan2}(\widehat{\Delta x},\widehat{\Delta z})\right)\right|.
 \label{eq:taskblind-bearing-residual}
\end{equation}
Residual units are meters and degrees; correctness against source targets is
measured separately.

The earlier capture-level bootstrap resamples 130 captures 10,000 times
(seed 20261011),
preserving each metric's micro-average in every draw. WorldIDE minus Base
query accuracy is $+1.07$ points, interval $[0.03,2.08]$; the point estimate
uses unrounded counts (7,752 versus 7,598 of 14,435 queries). The $+1.27$
video-macro gain uses a different estimand. Against Joint, the query gap is
$+0.55$ points, interval $[-0.44,1.54]$, and the distance-residual reduction is
0.060\,m, interval $[0.03083,0.08847]$\,m. These capture-level intervals
are sensitivity results; Section~\ref{app:state-joint-final} reports the
primary visit-level intervals used in the main text.
Against Base, distance residual decreases
by 0.076\,m, interval $[0.038,0.114]$\,m; the bearing-residual difference is
$-0.093^\circ$, interval $[-1.80,1.54]^\circ$. Entity-joint and endpoint-joint
gain intervals also include zero. These post-hoc intervals condition on fixed
heads and do not capture head-training variation.

\subsection{Frozen Cross-Operation Pair Diagnostics}
\label{app:pairacc-final}

The native diagnostic uses all 626 frozen source-selected pairs and 1,252
requests, with complete predictions for Base, Joint, WorldIDE, and the
reported ablations. Prompts, causal images, greedy 12-token decoding, parser,
and targets are shared. The operation request precedes the observation images,
so this protocol tests operation-conditioned native behavior; SIS prefix reuse
and the auxiliary probes use their separate protocols.

\paragraph{Matching conditions.}
Every pair matches entity category, endpoint block, and source-derived current
visibility. The 383 sector pairs also match the metric threshold and its
Boolean answer, while differing in signed sector. The 106 threshold pairs
match sector and threshold value while differing in its Boolean answer.
The 137 first-seen pairs match sector and a distance bin while differing in
first-seen block. Distance-bin boundaries are 0.5, 1, 2, 3, and 5 meters.
Neither a shared bin nor a shared threshold answer means equal true distance.
There are no independent last-seen, trajectory, or observer-motion pair groups
in this frozen manifest. Temporal targets retain the frustum-based visibility
proxy used in the original data.

\begin{table*}[t]
\centering\small
\caption{\textbf{Native distinguishability on all frozen pairs} (\%). Each model
cell reports single-history accuracy / PairAcc. Sources are captures / official
ARKit visits. Pooled accuracy weights all 626 pairs equally.}
\label{tab:pairacc-final}
\begin{tabular}{@{}lrrccccc@{}}
\toprule
Relation & Pairs & Sources & Base & Joint & WorldIDE & w/o trajectory & w/o categorical \\
\midrule
Signed sector & 383 & 128/117 & 22.85/3.39 & 26.11/4.18 & 23.11/3.13 & 23.37/4.70 & 26.89/5.48 \\
Distance threshold & 106 & 101/91 & 51.89/16.04 & 56.60/23.58 & 54.25/20.75 & 56.60/16.98 & 54.25/19.81 \\
First-seen block & 137 & 100/91 & 39.42/13.87 & 62.41/34.31 & 63.14/37.96 & 48.91/24.82 & 61.31/36.50 \\
Pooled & 626 & 130/119 & 31.39/7.83 & 39.22/14.06 & 37.14/13.74 & 34.58/11.18 & 39.06/14.70 \\
\bottomrule
\end{tabular}
\end{table*}

\paragraph{Paired uncertainty and semantic differences.}
The 130 captures map to 119 official ARKit visits. We resample visits jointly
across models 10,000 times (seed 20261012), preserving the pair-micro estimand.
Each cross-visit pair receives the product of its two visit multiplicities;
the nine within-visit pairs receive one multiplicity. No pair is removed.
This source-node bootstrap accounts for reuse of sources across pairs;
capture-level resampling is retained as a sensitivity analysis. Intervals are
pointwise and exploratory, without multiplicity correction. They condition on
the fixed source-selected pair graph and do not measure pair-construction
uncertainty or unseen-source generalization.

WorldIDE-minus-Base improves first-seen PairAcc by 24.09 points but changes
sector PairAcc by $-0.26$. The directly paired difference between these two
gains is 24.35 points, interval $[8.73,41.91]$. Joint shows the same temporal
emphasis. WorldIDE-minus-Joint intervals include zero in every semantic group
and in the pooled comparison (Table~\ref{tab:pairacc-deltas-final}).

\begin{table*}[t]
\centering\small
\caption{\textbf{Paired native PairAcc differences} in percentage points, with
95\% visit-bootstrap intervals. Every model contrast uses identical pairs.}
\label{tab:pairacc-deltas-final}
\begin{tabular}{@{}llrr@{}}
\toprule
Relation & Contrast & Difference & 95\% interval \\
\midrule
Signed sector & Joint--Base & +0.78 & $[-3.41,5.43]$ \\
Signed sector & WorldIDE--Base & -0.26 & $[-3.88,3.19]$ \\
Signed sector & WorldIDE--Joint & -1.04 & $[-4.71,2.00]$ \\
Signed sector & WorldIDE--w/o trajectory & -1.57 & $[-5.06,1.31]$ \\
Signed sector & WorldIDE--w/o categorical & -2.35 & $[-6.49,1.33]$ \\
Distance threshold & Joint--Base & +7.55 & $[-6.38,23.03]$ \\
Distance threshold & WorldIDE--Base & +4.72 & $[-5.32,16.05]$ \\
Distance threshold & WorldIDE--Joint & -2.83 & $[-14.95,9.60]$ \\
Distance threshold & WorldIDE--w/o trajectory & +3.77 & $[-11.32,17.44]$ \\
Distance threshold & WorldIDE--w/o categorical & +0.94 & $[-8.99,11.29]$ \\
First-seen block & Joint--Base & +20.44 & $[4.91,37.86]$ \\
First-seen block & WorldIDE--Base & +24.09 & $[9.01,41.03]$ \\
First-seen block & WorldIDE--Joint & +3.65 & $[-6.50,15.44]$ \\
First-seen block & WorldIDE--w/o trajectory & +13.14 & $[0.00,27.91]$ \\
First-seen block & WorldIDE--w/o categorical & +1.46 & $[-8.33,11.11]$ \\
Pooled & Joint--Base & +6.23 & $[1.28,11.43]$ \\
Pooled & WorldIDE--Base & +5.91 & $[1.29,10.94]$ \\
Pooled & WorldIDE--Joint & -0.32 & $[-4.34,3.58]$ \\
Pooled & WorldIDE--w/o trajectory & +2.56 & $[-1.91,6.83]$ \\
Pooled & WorldIDE--w/o categorical & -0.96 & $[-4.65,2.39]$ \\
\bottomrule
\end{tabular}
\end{table*}

\paragraph{Direct family-removal inference.}
We additionally evaluate the existing without-trajectory and without-categorical
checkpoints on all 1,252 frozen requests, yielding 626 complete pairs per model.
Inference uses the existing benchmark ablation models.
The original eight-way shard membership, batch size two, image policy,
200,704-pixel ceiling, greedy 12-token decoding, and parser are preserved.
Inputs are materialized on CPU with the original encoding function before GPU
inference; the two models have identical processor artifacts. These controls
retain the benchmark ablation checkpoints and their fixed training budgets.

WorldIDE-minus-no-trajectory first-seen gain is 13.14 points, interval
$[0.00,27.91]$, while its sector gain is $-1.57$, interval $[-5.06,1.31]$.
The directly paired temporal-minus-sector interaction is 14.71 points,
interval $[1.06,29.77]$. The corresponding categorical-removal interaction
is 3.81 points, interval $[-6.52,14.18]$. WorldIDE and the no-trajectory
control produce parse-valid outputs for every request. The latter has three
future-block predictions versus zero for WorldIDE, compared with 18 additional
correct temporal pairs for WorldIDE; format or boundary repair alone cannot
account for the complete temporal difference. The first-seen interval touches zero, while the interaction interval is
positive. These contrasts characterize supervision-pool composition under
fixed updates, including changes in source coverage and remaining-example
exposure. Pair inference is unavailable for observer-motion removal; the
located metric model could not be bound to the reported metric ablation.

\paragraph{Failure patterns and readout scope.}
On sector pairs, WorldIDE succeeds on 12/383 pairs versus 13/383 for Base.
It gives identical predictions to the two different-target histories on
145 pairs (Base: 127), and two different but incorrect predictions on
158 pairs (Base: 154). First-seen successes increase from 19/137 to 52/137.
Base also has 11 first-seen pairs with an invalid parsed output, compared
with zero for the trained models; the 33 additional correct pairs therefore
cannot all be attributed to format repair.

Using the existing auxiliary heads on the identical pair IDs gives pooled
PairAcc of 15.81/13.10/13.74 for Base/Joint/WorldIDE. First-seen auxiliary
PairAcc is 20.44/15.33/18.98, while sector PairAcc is 14.36/12.53/12.79.
The auxiliary first-seen scores differ from the native improvement pattern.
The two interfaces use different encoding and reader-training protocols.

\subsection{Joint Recovery and Geometric Correctness}
\label{app:state-joint-final}

The task-blind analysis retains all 14,435 queries, 1,445 eligible entities,
and 514 endpoints. We add five fixed semantic combinations, each scored per
entity with its original source target and scoring rule. These exploratory
conjunctions supplement the full endpoint criterion. Distance and
threshold, or bearing and sector, share a predicted field and are not counted
as independent capabilities in a two-operation conjunction.

Entity labels come from source OBB category annotations. An entity is eligible
once registered by the geometric visibility proxy; categories with multiple
eligible objects at an endpoint are excluded. The category index enters a
64-dimensional embedding at the head input, while the encoder receives no
entity label or query. Source UIDs are provenance only. This supplies referent
identity and eligibility, but no target coordinates, bearing, or event turns.
The protocol measures recoverability given source-resolved entities.

The primary paired bootstrap resamples the 119 official visits containing
the 130 captures, using 10,000 draws and seed 20261012. All models share
each draw, with ratio-of-sums resampling appropriate to each metric. Capture
intervals above are sensitivity results. For example, the WorldIDE-minus-Base
query gain remains $+1.07$ points, but its visit interval is $[-0.13,2.20]$;
WorldIDE-minus-Joint is $+0.55$, interval $[-0.54,1.60]$.

\begin{table*}[t]
\centering\small
\caption{\textbf{Exploratory two-operation recovery} (\%; 1,445 entities).
Each cell gives the two marginal accuracies followed by their joint accuracy.
All WorldIDE-minus-Base and WorldIDE-minus-Joint joint intervals include zero.}
\label{tab:joint-combinations-final}
\begin{tabular}{@{}lccc@{}}
\toprule
Operation pair & Base: marginals / joint & Joint: marginals / joint & WorldIDE: marginals / joint \\
\midrule
Distance + bearing & 23.94, 17.23 / 4.71 & 24.08, 17.79 / 5.05 & 25.19, 18.27 / 4.84 \\
Distance + sector & 23.94, 43.46 / 12.18 & 24.08, 46.71 / 12.66 & 25.19, 49.07 / 13.15 \\
First + last seen & 81.25, 63.18 / 54.19 & 78.20, 65.26 / 53.43 & 77.85, 65.95 / 53.49 \\
Distance + first seen & 23.94, 81.25 / 18.55 & 24.08, 78.20 / 18.34 & 25.19, 77.85 / 19.93 \\
Bearing + first seen & 17.23, 81.25 / 14.19 & 17.79, 78.20 / 13.43 & 18.27, 77.85 / 14.39 \\
\bottomrule\end{tabular}\end{table*}

\paragraph{Geometric reference and dimensions.}
For the task-blind probe, every entity is represented by the centroid of
its source-annotated oriented bounding box. The same endpoint pose transforms
this point into camera coordinates. Right and forward components are measured
in meters, and their source distance is $d=\sqrt{\Delta x^2+\Delta z^2}$.
This is the projection onto the camera's right--forward plane, with the
camera-vertical component omitted; it is neither a three-dimensional norm
nor a distance to the object's nearest surface. The 1,445 validation records
satisfy this target identity to floating-point precision. The camera-to-TV
nearest-point training example in Section~\ref{sec:relational-learning}
uses a separate operation definition and does not supply these probe targets.

\paragraph{Agreement versus accuracy.}
Table~\ref{tab:state-recoverability} reports $e_d$, the arithmetic mean of
$r_d(h,e)$ in Eq.~\ref{eq:distance-compatibility} over all 1,445 eligible
entity--endpoint records, with equal record weight and units of meters.
Each record's fields refer to the same entity, time, and endpoint camera frame.
Distance-residual means are 0.621/0.605/0.546\,m for Base/Joint/WorldIDE.
Source-distance MAEs are 0.790/0.800/0.785\,m. WorldIDE reduces the residual
relative to Joint by 0.05969\,m, with paired 95\% visit-bootstrap interval
$[0.03139,0.08765]$\,m; these round to the main text's 0.060\,m and
$[0.031,0.088]$\,m. Its source-distance error difference is $-0.015$\,m,
interval $[-0.043,0.013]$. Among the 799
entities with a lower residual than Joint, only 410 also have lower
source-distance error; 389 do not. A residual at most 0.25\,m coupled with
source-distance error greater than 0.25\,m occurs for 20.62\% of WorldIDE
entities versus 15.57\% for Joint, a 5.05-point increase with interval
$[2.11,8.02]$. This exploratory diagnostic uses the existing metric tolerance as its
reference threshold.

Source-bearing mean circular error is 63.31/60.79/58.99 degrees.
WorldIDE-minus-Base is $-4.32^\circ$, interval $[-6.63,-1.98]^\circ$;
WorldIDE-minus-Joint is $-1.80^\circ$, interval $[-4.09,0.51]^\circ$.
Thus some source-grounded angular information improves even though the
native sector-pair diagnostic does not. The scores characterize different
operations, populations, and fitted readout interfaces.

The full endpoint successes remain 1/514, 0/514, and 2/514, and entity-level
seven-operation successes remain 26/1,445, 34/1,445, and 34/1,445.
The accompanying evidence tables retain
point estimates, paired intervals, operation counts, source IDs, and cases.

\section{Supplementary Diagnostics}
\label{app:diagnostic-boundaries}

\stitle{Independent cross-operator control.}
On 32 source-disjoint videos, the WorldIDE checkpoint is evaluated on
256 positive/negative compatibility judgments and 128 direct operator queries.
Base obtains 52.73\% compatibility and 33.59\% direct accuracy; WorldIDE obtains
51.17\% and 35.16\%. The evaluation criterion requires at least two points of improvement on
both metrics and is unmet.

\stitle{Planning boundary.}
SIS perceptual self-awareness and memory improve, while self-awareness
reasoning falls from 51.5 to 41.9 and Path Planning falls from 45.6 to 27.9 over
272 questions. The same pattern appears in the matched joint control and two
independent SIS checkpoints (27.2--30.9). Observed-state supervision therefore does not
establish future action-policy reasoning.

\stitle{Archived successful readout.}
At endpoint \texttt{778ba127\ldots::3}, all six available queries are correct.
The State Plane predicts distance $0.78$\,m for a $0.61$\,m target and bearing
$88.08^\circ$ for an $83.73^\circ$ target; sector, threshold, entity, and turn
readouts are also correct. Replacing the state changes the distance to
$1.41$\,m and the entity class from the correct class 15 to class 1. This case
shows one endpoint supporting several compatible operations while retaining
source-sensitive quantities.

\stitle{Archived failure case.}
At endpoint \texttt{e01dd61c\ldots::3}, entity and turn are correct, but metric
and directional readouts fail together: predicted versus target distance is
$0.71$ versus $1.10$\,m, bearing is $108.60^\circ$ versus $152.26^\circ$, and
the derived sector is left rather than behind. The threshold query remains
correct. This case illustrates why local operator success and schema-valid
execution do not imply a fully coherent geometric state. Full case identities
and predictions remain in the archived query-execution receipt.

\section{Training and Evaluation Summary}
\label{app:evaluation}

\stitle{Training and controls.}
WorldIDE and its matched controls start from Qwen3-VL-8B-Instruct and use the same source-grounded records, optimization setup, and training budget. The controlled comparisons vary the supervision specified in Appendix~\ref{app:relational-training}.

\stitle{Benchmark evaluation.}
All variants use the official OST, UCS, and SIS test populations and scoring protocols under matched benchmark-specific observation and decoding interfaces. The few SIS questions affected by a corrupt released clip use the same recovered evidence for every comparison arm. The headline mean weights the three benchmarks equally, and score differences are computed before rounding.

\stitle{Functional probes.}
State Plane heads are fitted independently on frozen encoders with matched probe settings. Paired-history and geometric uncertainty use source-clustered resampling and describe variation across sources for the fitted models.

\section{Qualitative Comparisons with Published Baselines}
\label{app:directme-case}

\stitle{Comparison scope.}
These selected qualitative cases compare archived predictions under their
published or recorded protocols. Model size, observation access, and compute
are method-specific. We describe observable answers and retrieval traces;
aggregate performance is evaluated by the matched benchmarks in the main
text. Complete inventories retain reverse outcomes and alignment exclusions.

\subsection{VideoChat3, JoyAI, and Mage-VL on the Same Questions}
\label{app:multimethod-cases}

We compare archived \method{} predictions with recorded outputs from
VideoChat3-4B, JoyAI-VL-Interaction, and Mage-VL on 19 questions from three
pre-existing manuscript scenes: seven OST, six UCS, and six SIS questions.
All three external predictions and the reference are retained for every
question. The six representative figures below cover first-appearance order,
temporal co-occurrence, counting, door-state recall, preceding-landmark
memory, and flight-action recall. Each displays all four methods, including
correct baseline predictions, plus concise differences from the reference. In the OST Turn-5 example, all three external
models miss the reference turn, whereas \method{} returns Turn 2.

\stitle{Inputs and interpretation.}
The external visual packets contain the complete causal OST history through
the query turn, uniform causal UCS frames through the question time, or the
official 32-frame SIS packet. Each figure combines four overview observations, two labelled detail views,
and a compact table of all archived questions from the same scene.
The highlighted table row is the focal question. Additional query rows provide
context and are not extra causal input to an earlier query. The SIS table also
shows WorldIDE's recorded reused-prefix and other-video-prefix outputs;
the UCS panels include the explicitly identified DirectMe author-demo
retrieval counts. The UCS protocol requests 128 frames, but shorter prefixes contain fewer
unique source frames; displayed indices refer to the actual packet images.

\stitle{Case inventory.}
Ten of the 19 questions have a correct \method{} answer and at least one
incorrect external answer. Two reverse cases are retained: OST Turn 7 (VideoChat3 correct, WorldIDE incorrect) and the
UCS sofa count at 12\,s (Mage-VL correct, WorldIDE incorrect).
The distance question with unassigned WorldIDE correctness remains unscored.
The complete 12-page figure collection contains all ten positive contrasts
and both reverse cases; the first six are the representative set below.

\begin{figure*}[p]
\centering
\includegraphics[width=\textwidth,height=0.91\textheight,keepaspectratio]{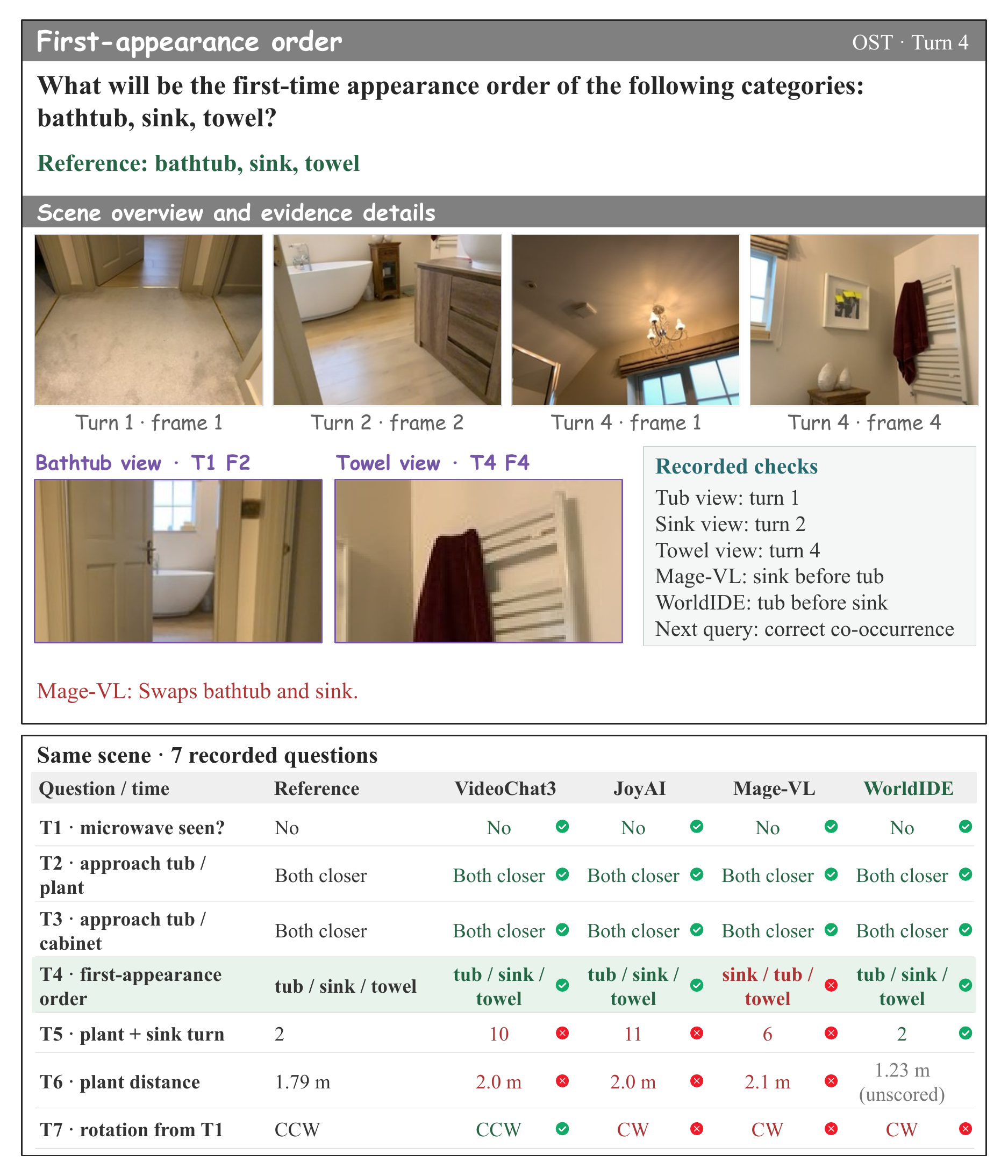}
\caption{\textbf{First-appearance order} (OST, Turn 4).}
\label{fig:multimethod-1}
\end{figure*}

\begin{figure*}[p]
\centering
\includegraphics[width=\textwidth,height=0.91\textheight,keepaspectratio]{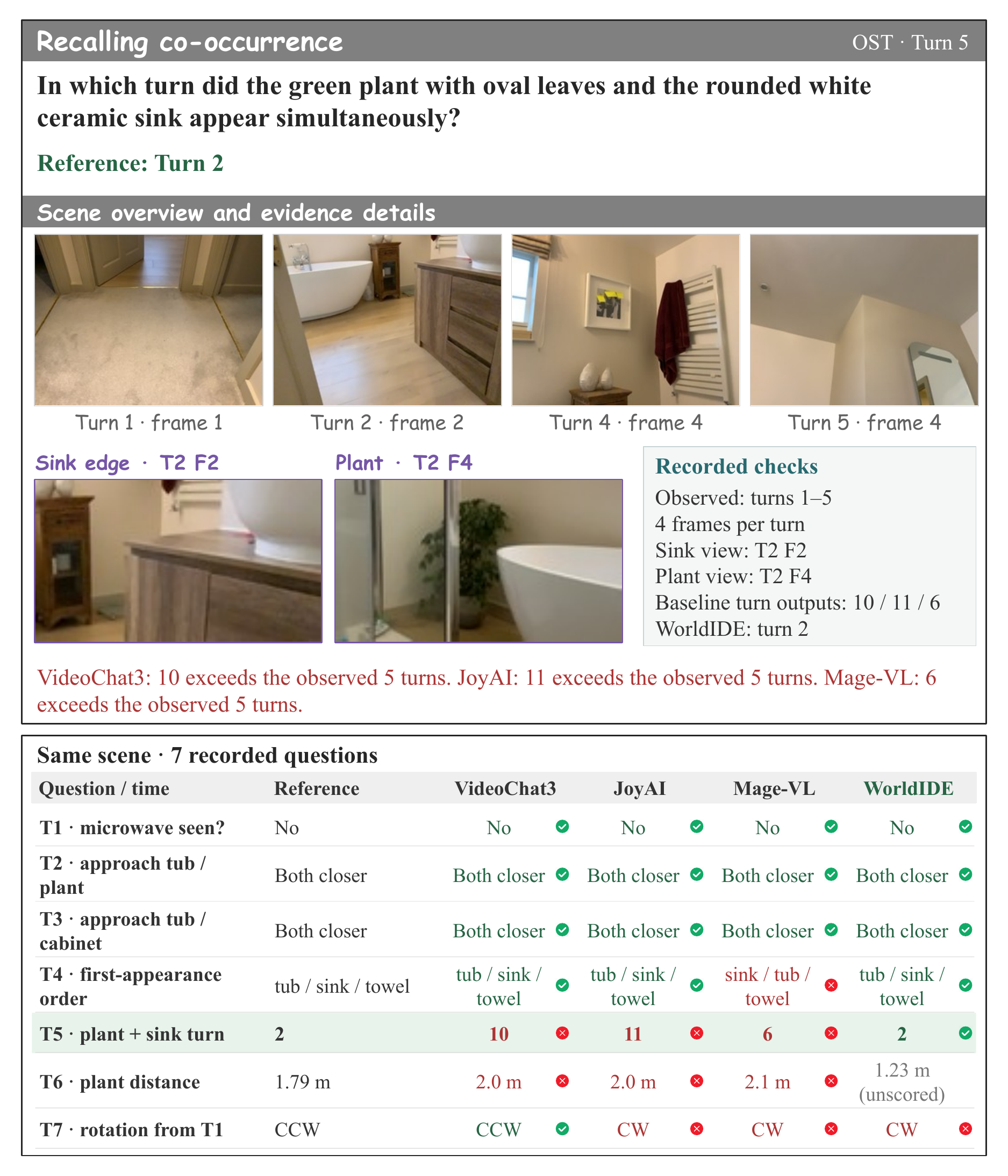}
\caption{\textbf{Recalling co-occurrence} (OST, Turn 5).}
\label{fig:multimethod-2}
\end{figure*}

\begin{figure*}[p]
\centering
\includegraphics[width=\textwidth,height=0.91\textheight,keepaspectratio]{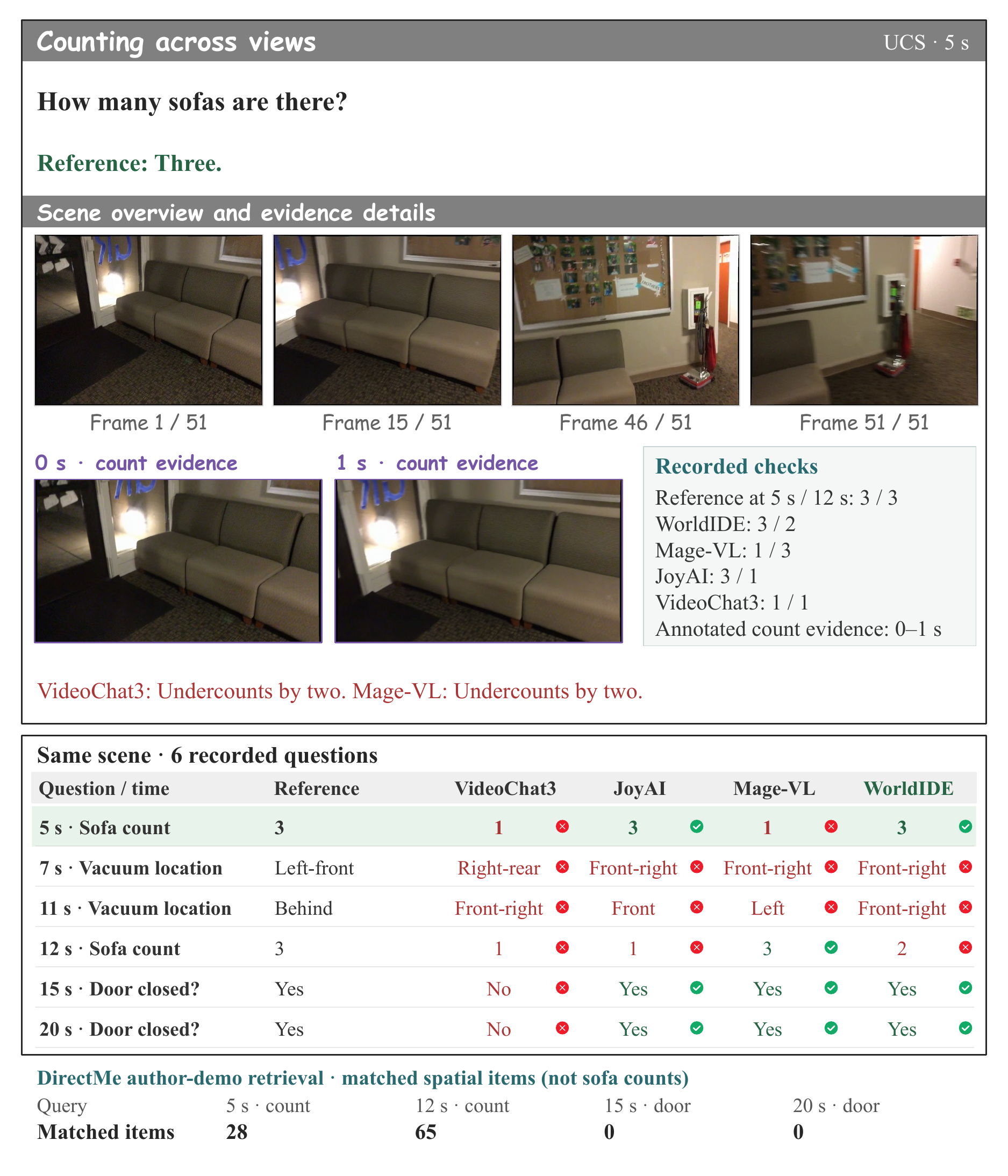}
\caption{\textbf{Counting across views} (UCS, 5 s).}
\label{fig:multimethod-3}
\end{figure*}

\begin{figure*}[p]
\centering
\includegraphics[width=\textwidth,height=0.91\textheight,keepaspectratio]{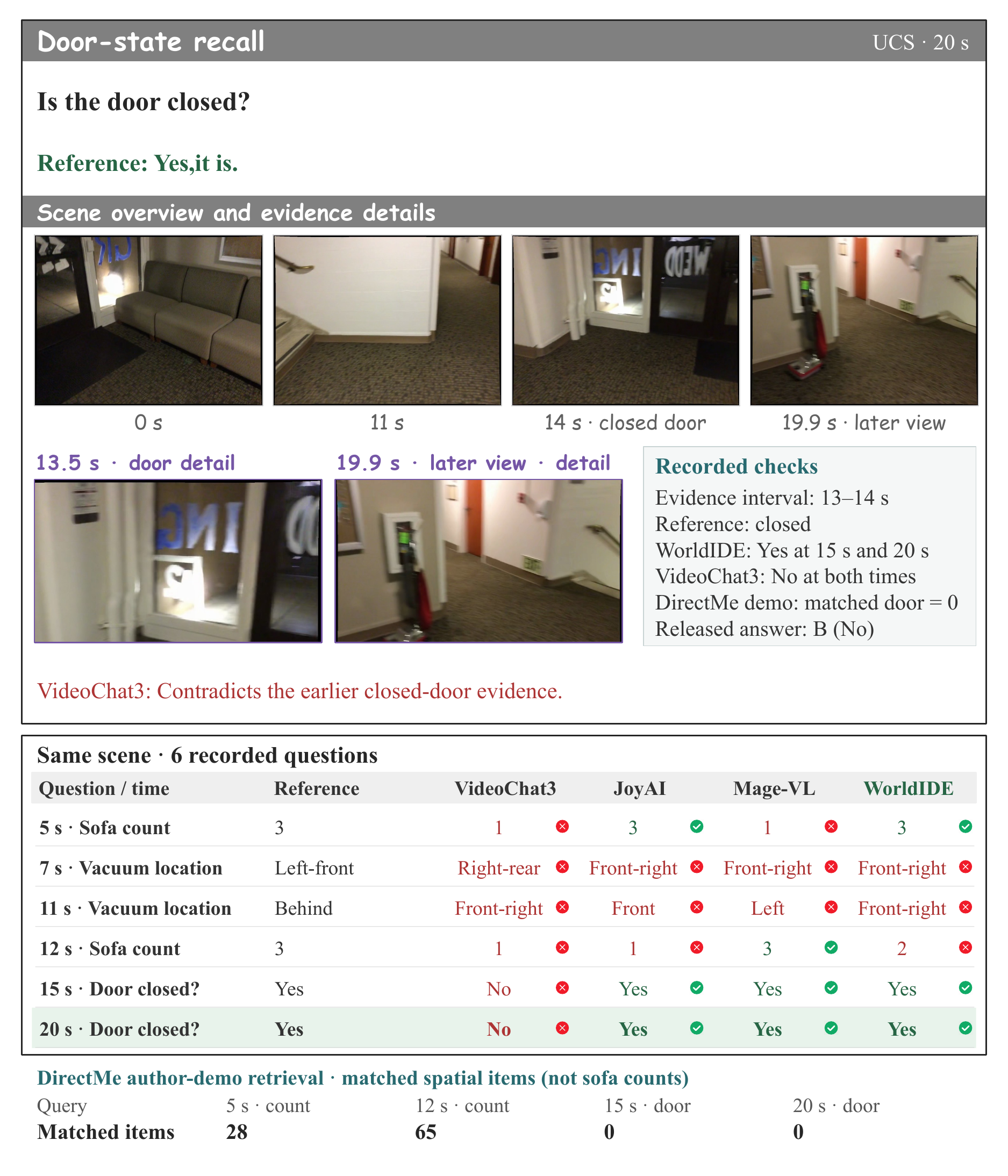}
\caption{\textbf{Door-state recall} (UCS, 20 s).}
\label{fig:multimethod-4}
\end{figure*}

\begin{figure*}[p]
\centering
\includegraphics[width=\textwidth,height=0.91\textheight,keepaspectratio]{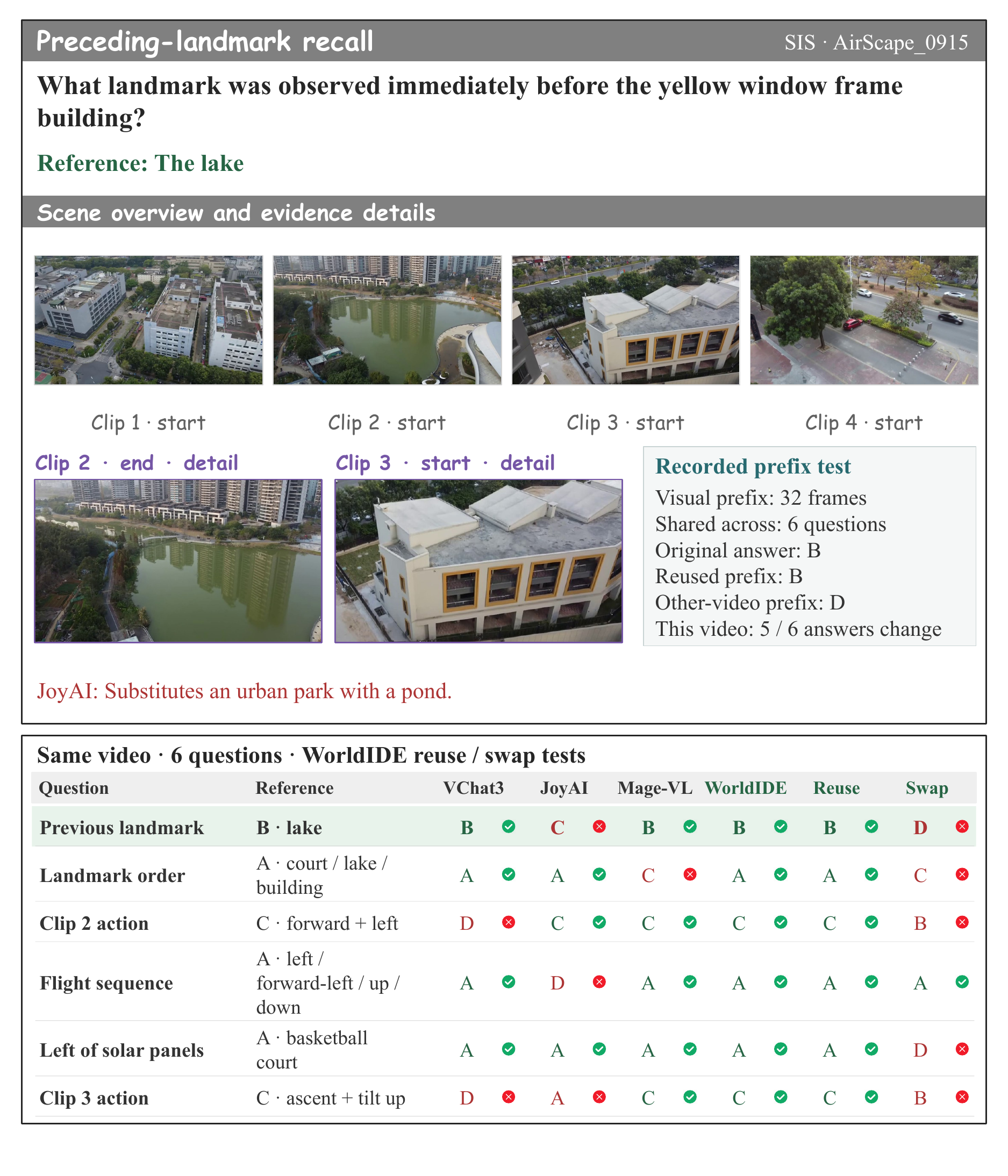}
\caption{\textbf{Preceding-landmark recall} (SIS).}
\label{fig:multimethod-5}
\end{figure*}

\begin{figure*}[p]
\centering
\includegraphics[width=\textwidth,height=0.91\textheight,keepaspectratio]{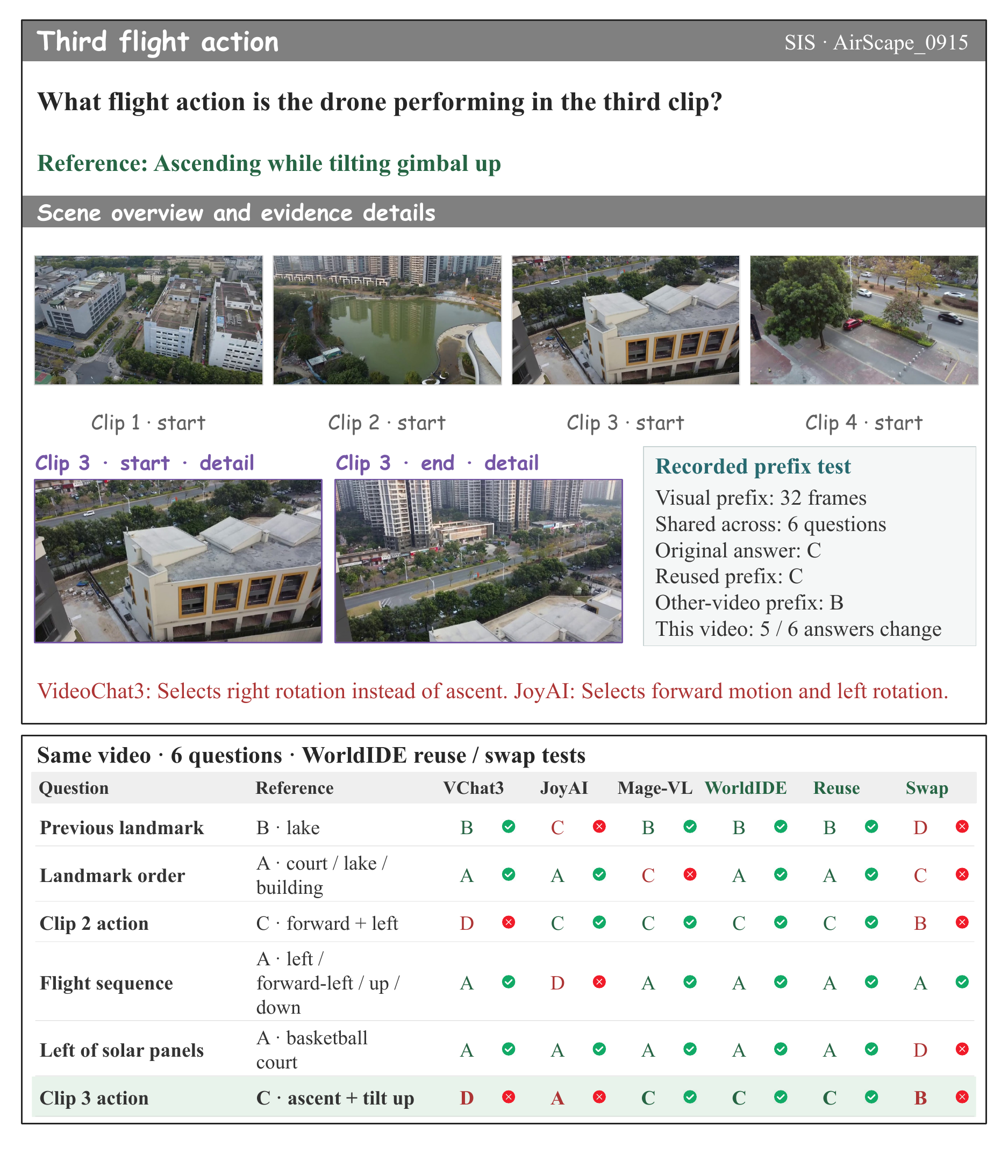}
\caption{\textbf{Third flight action} (SIS).}
\label{fig:multimethod-6}
\end{figure*}

\clearpage
\subsection{DirectMe: Historical Door-State Recall}

\begin{figure*}[p]
\centering
\includegraphics[width=0.98\textwidth,height=0.76\textheight,keepaspectratio]{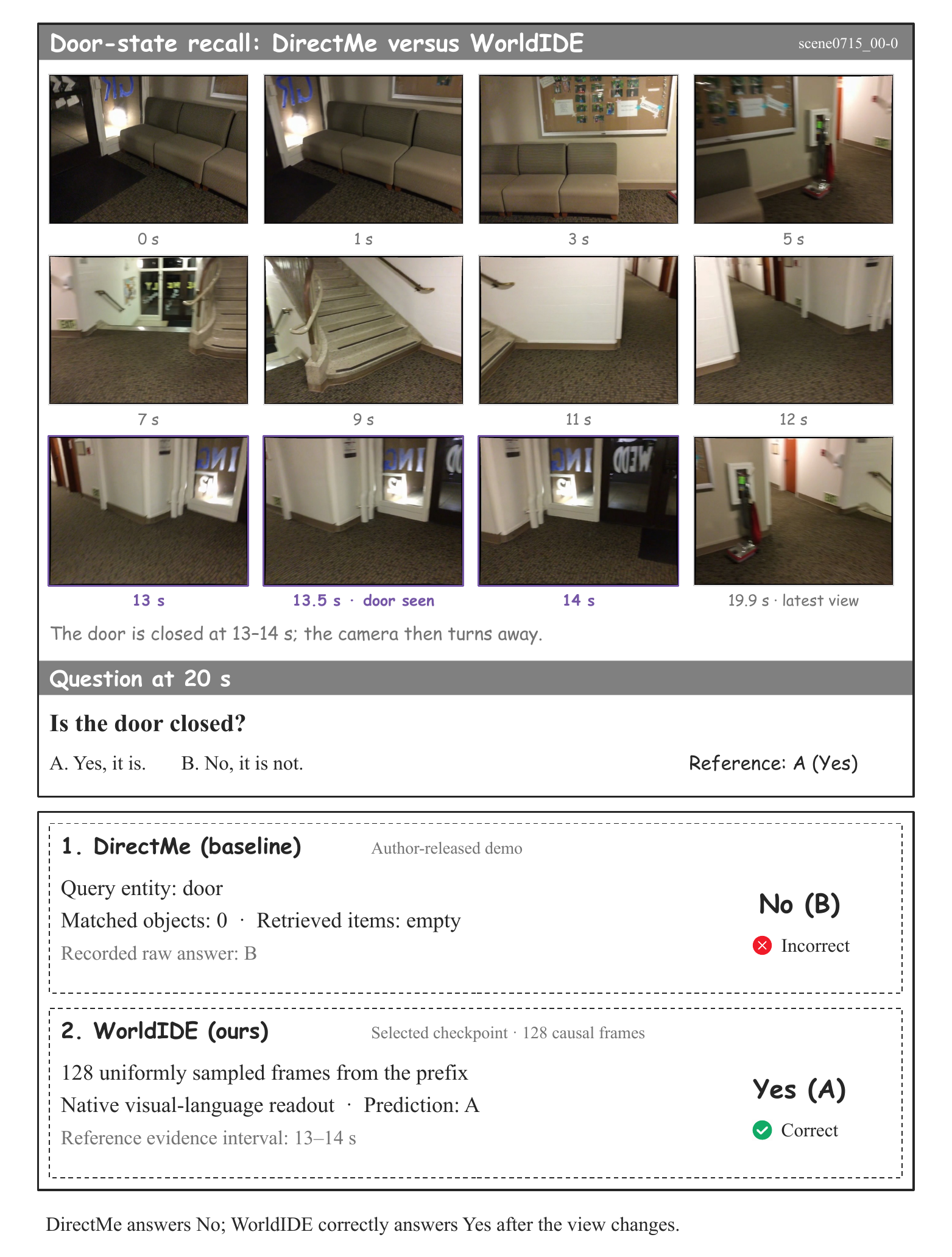}
\caption{\textbf{DirectMe versus WorldIDE: a concrete historical door-state
query.} At 20\,s, DirectMe's author-released trace has no matching door object
and answers No; WorldIDE answers Yes, matching the reference and the earlier
closed-door observation. Frames are selected from the same original video and
labelled by time. The complete released-record comparison appears in
Table~\ref{tab:directme-release-overlap}.}
\label{fig:case-directme}
\end{figure*}

Figure~\ref{fig:case-directme} compares \method{} with the external spatial
memory method DirectMe~\citep{wang2026ucs} on a concrete UCS question.
DirectMe's answer and retrieval trace come from the authors' released
demonstration outputs; WorldIDE uses its archived final-checkpoint prediction.
These demo outputs are not the per-question outputs underlying the
published 50.5\% benchmark row. Its exact backbone, quantization, and frame budget are unreported.

\stitle{The same query and reference.}
The case is \texttt{scene0715\_00-0\_4}, at 20\,s: ``Is the door closed?''
Both records use reference label A, corresponding to Yes; option B is No.
The official evidence annotation identifies the door as closed at 13--14\,s.
The final view faces away from that door, making the earlier observation
relevant to the later question. Figure~\ref{fig:case-directme} shows selected original frames with their times.

\stitle{DirectMe's recorded retrieval and answer.}
DirectMe parses the entity label \emph{door}, but reports zero matched objects
and an empty list of retrieved spatial items. Its recorded raw answer is B
(No), marked incorrect in the released artifact. The trace records no retrieved door evidence in this example.

\stitle{WorldIDE's recorded answer.}
The final WorldIDE checkpoint predicts A (Yes), which matches the reference.
The record specifies 128 uniformly sampled causal frames, native letter
parsing, and no additional retrieval request. Thus WorldIDE produces the
correct historical door-state answer in a case where the external method's
released retrieval trace contains no matching door object.

\stitle{Complete released-record overlap.}
We compare all seven released DirectMe records with the available
WorldIDE outputs and official question metadata. Six question identifiers
occur in the current WorldIDE evaluation; reference option labels agree for
five. The 15\,s door query has reference A in the DirectMe artifact but B in
the current question file, and is excluded from letter-level comparison. The
photo-location query is absent from the current evaluation and is not counted
as a WorldIDE error. Among the five reference-label-aligned records, one favors
WorldIDE, two favor DirectMe, one is a shared success, and one is a shared
failure. Table~\ref{tab:directme-release-overlap} retains these reverse outcomes and
alignment exclusions.

\begin{table}[t]
\centering
\footnotesize
\setlength{\tabcolsep}{3pt}
\begin{tabularx}{\columnwidth}{Xll}
\toprule
Query (time) & DirectMe & WorldIDE \\
\midrule
Sofas (5\,s) & Correct & Correct \\
Vacuum (7\,s) & Incorrect & Incorrect \\
Vacuum (11\,s) & Correct & Incorrect \\
Sofas (12\,s) & Correct & Incorrect \\
Door (15\,s)$^\dagger$ & Incorrect & Correct \\
\rowcolor{MethodBlue}Door (20\,s) & Incorrect & Correct \\
Photos (20\,s)$^\ddagger$ & Correct & -- \\
\bottomrule
\end{tabularx}
\caption{\textbf{Complete released-record overlap.} Each column uses its
original correctness flag. $^\dagger$Reference labels conflict (A versus B),
so this row is excluded from paired comparison. $^\ddagger$No corresponding
WorldIDE record; unscored. The table includes all released records.}
\label{tab:directme-release-overlap}
\end{table}

\subsection{Published SIS Baseline versus WorldIDE}
\label{app:published-sis-cases}

\begin{figure*}[p]
\centering
\includegraphics[width=0.98\textwidth,height=0.79\textheight,keepaspectratio]{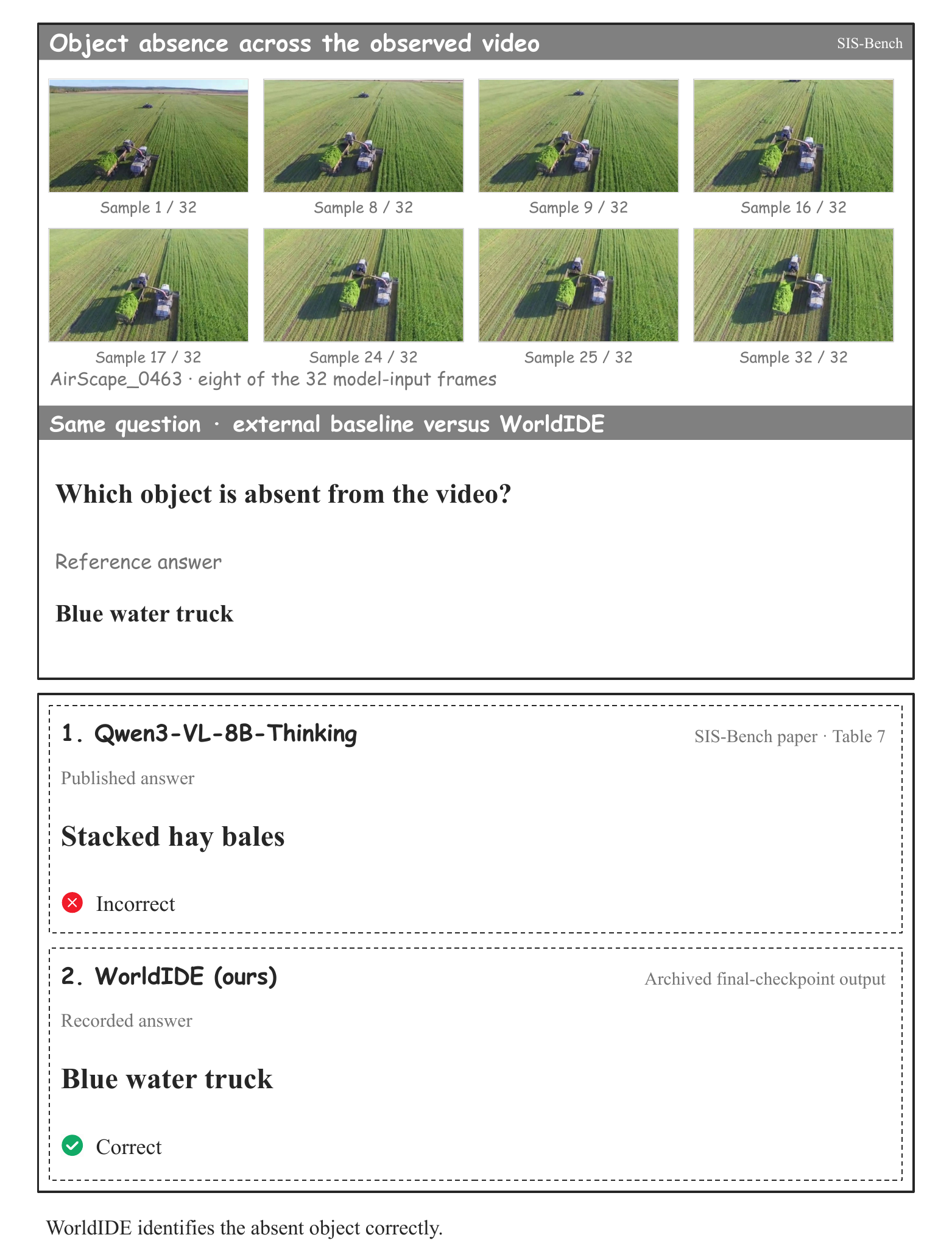}
\caption{\textbf{Object absence: published baseline versus WorldIDE.} The published baseline selects stacked hay bales as absent. WorldIDE selects the blue water truck, matching the reference. }
\label{fig:external-absence}
\end{figure*}

\begin{figure*}[p]
\centering
\includegraphics[width=0.98\textwidth,height=0.79\textheight,keepaspectratio]{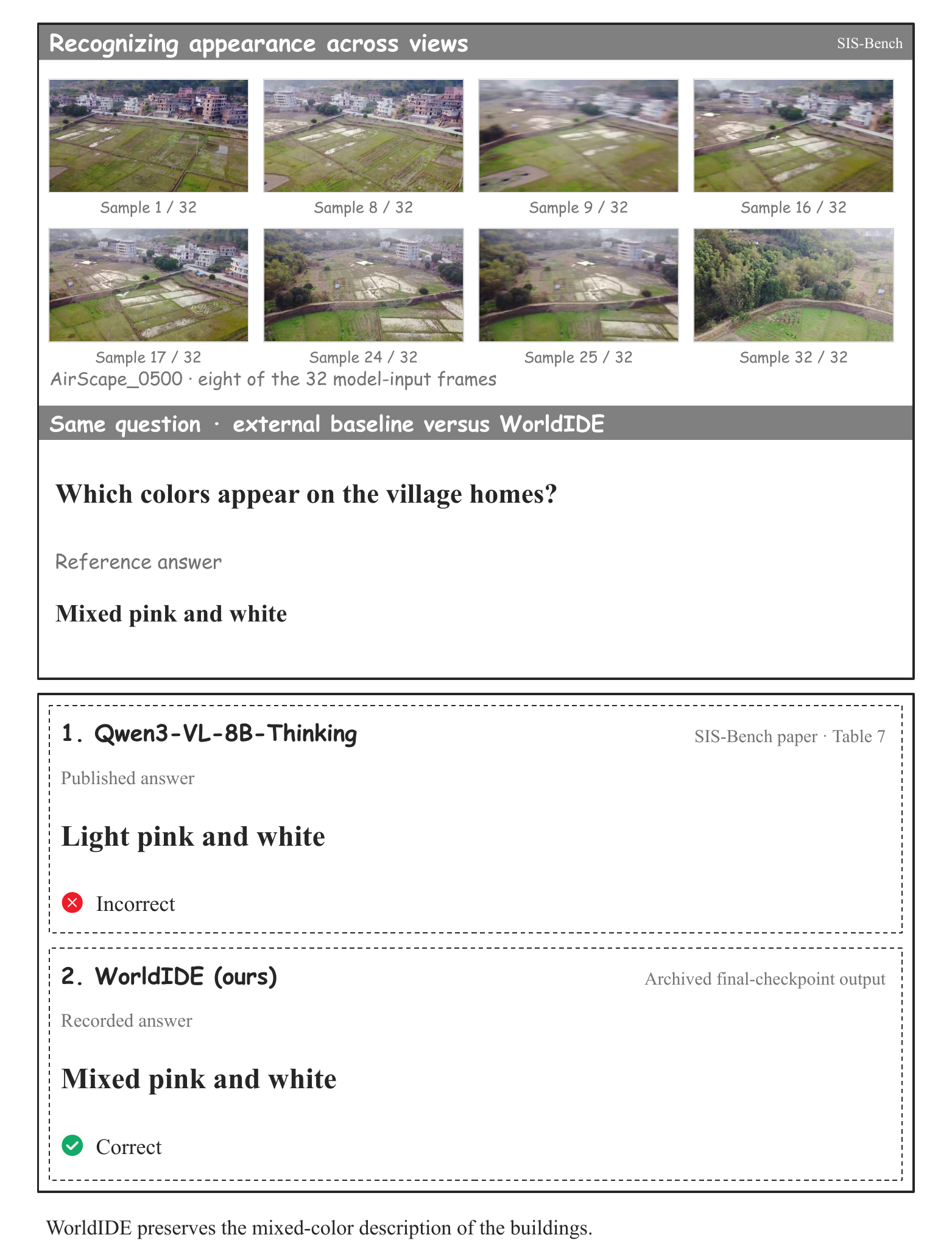}
\caption{\textbf{Building appearance: published baseline versus WorldIDE.} The published baseline selects light pink and white. WorldIDE selects mixed pink and white, matching the reference. }
\label{fig:external-appearance}
\end{figure*}

\begin{figure*}[p]
\centering
\includegraphics[width=0.98\textwidth,height=0.79\textheight,keepaspectratio]{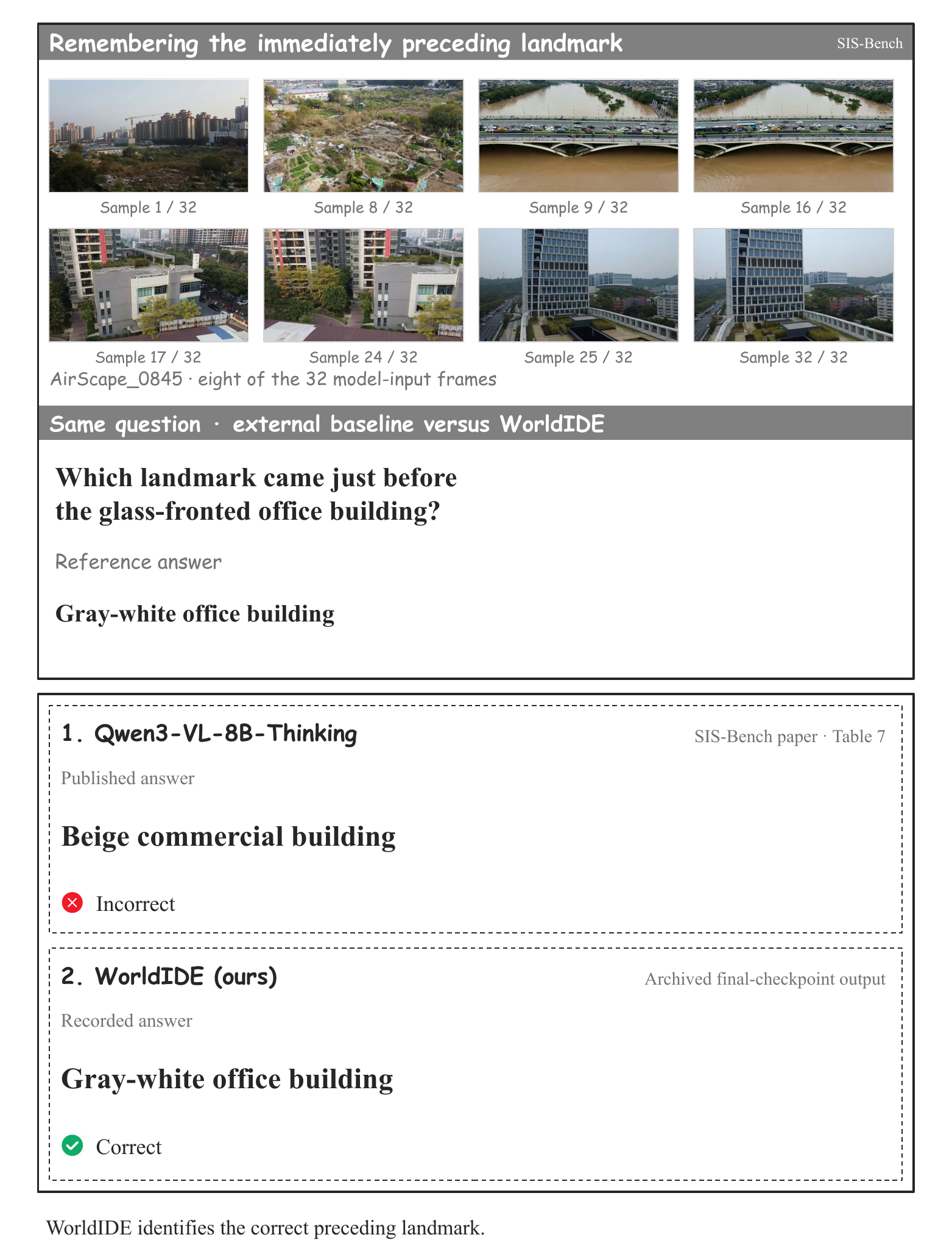}
\caption{\textbf{Preceding landmark recall: published baseline versus WorldIDE.} The published baseline selects a beige commercial building. WorldIDE selects the gray-white office building immediately preceding the queried landmark. }
\label{fig:external-recall}
\end{figure*}

\begin{figure*}[p]
\centering
\includegraphics[width=0.98\textwidth,height=0.79\textheight,keepaspectratio]{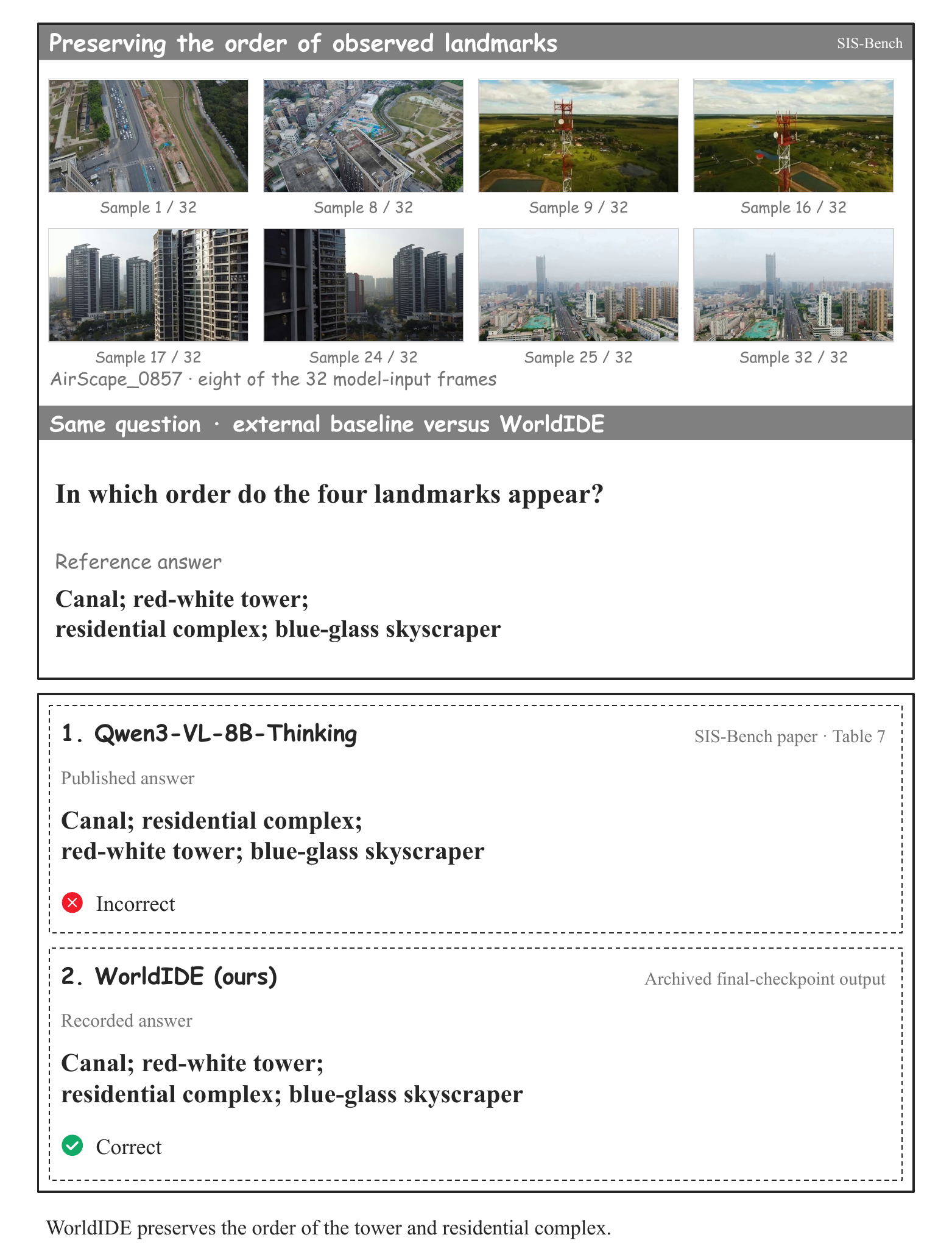}
\caption{\textbf{Landmark order: published baseline versus WorldIDE.} The published baseline swaps the communication tower and residential complex. WorldIDE preserves their reference order. }
\label{fig:external-order}
\end{figure*}

Figures~\ref{fig:external-absence}--\ref{fig:external-order} compare
\method{} with the Qwen3-VL-8B-Thinking baseline reported in Table~7 of
\citet{zou2026sis}. This is the published reasoning-model baseline, distinct
from the Qwen3-VL-8B-Instruct initialization used by \method{}.
The four selected questions concern object absence, building appearance,
immediately preceding landmark recall, and temporal landmark order.
The published baseline answer is incorrect and the archived \method{} answer
is correct in each selected comparison.

\stitle{Question and answer alignment.}
The source table omits question and video identifiers and abbreviates the
questions. We align each case using its distinctive question referents,
answer choices, and reference meaning in the current dataset. The selected
current identifiers are \texttt{object\_existence\_0036},
\texttt{object\_attribute\_0012}, \texttt{landmark\_recall\_0001}, and
\texttt{landmark\_appearance\_order\_0036}. Option letters are reordered
between the publication and current dataset, so the figures compare answer
text rather than raw cross-release letters. Eight of the 32 archived-protocol
input frames are displayed for each video.

The publication supplies summarized baseline answers. Examples lacking a
unique alignment, or also missed by \method{}, are excluded from the set of
corrected cases.

\clearpage
\begin{figure*}[p]
\subsection{Supplementary Failure Cases}
The following cases illustrate counting and cross-query failures.\par\medskip
\centering
\includegraphics[width=0.98\textwidth,height=0.79\textheight,keepaspectratio]{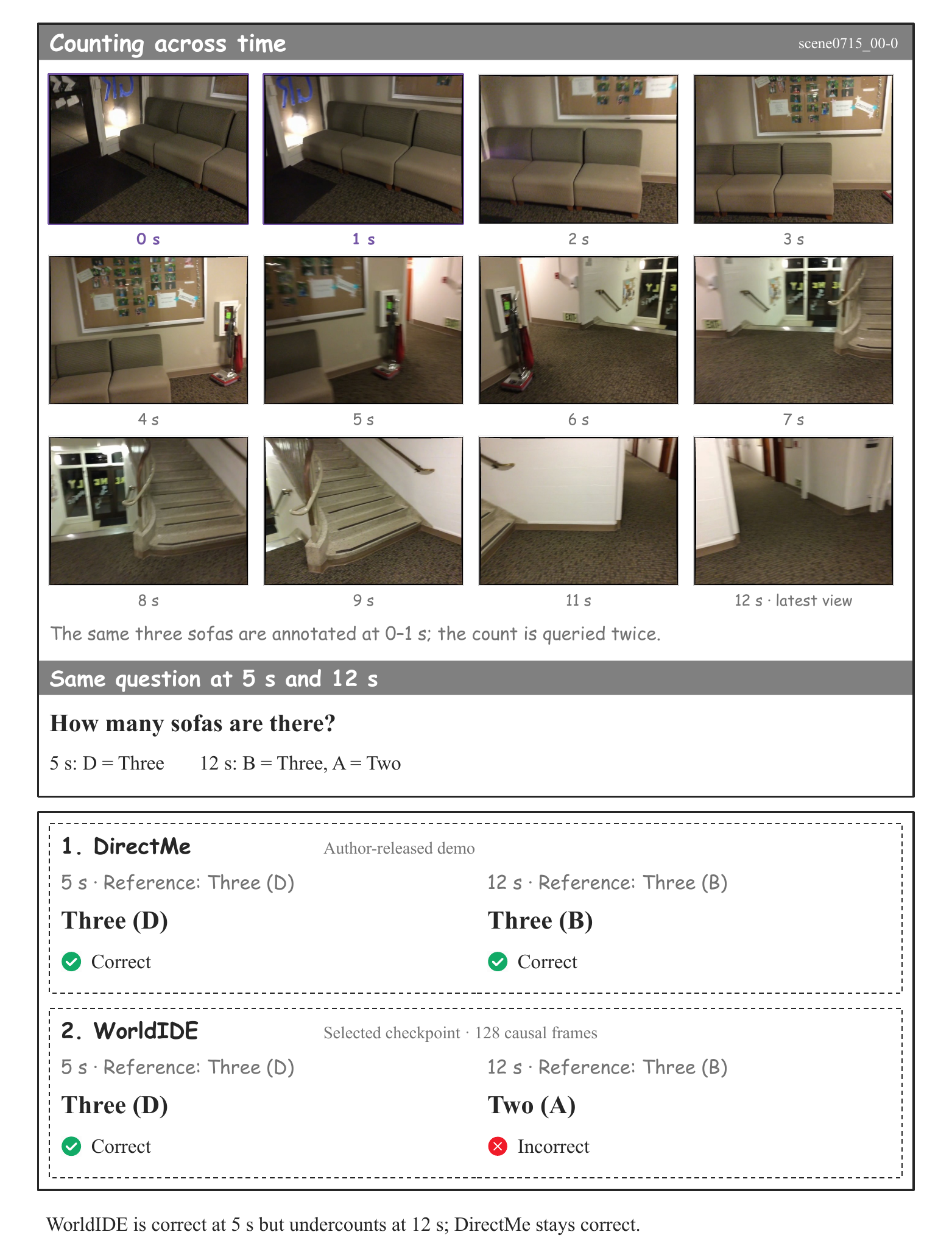}
\caption{\textbf{Counting across time in the same video.} The reference count is three sofas at both 5\,s and 12\,s. DirectMe answers correctly at both times; WorldIDE changes from the correct three to an incorrect two. Option letters differ between timestamps and are shown with their answer meanings. Selected original frames illustrate the shared scene; each prediction uses only its own causal prefix. }
\label{fig:case-directme-count}
\end{figure*}

\begin{figure*}[p]
\centering
\includegraphics[width=0.98\textwidth,height=0.79\textheight,keepaspectratio]{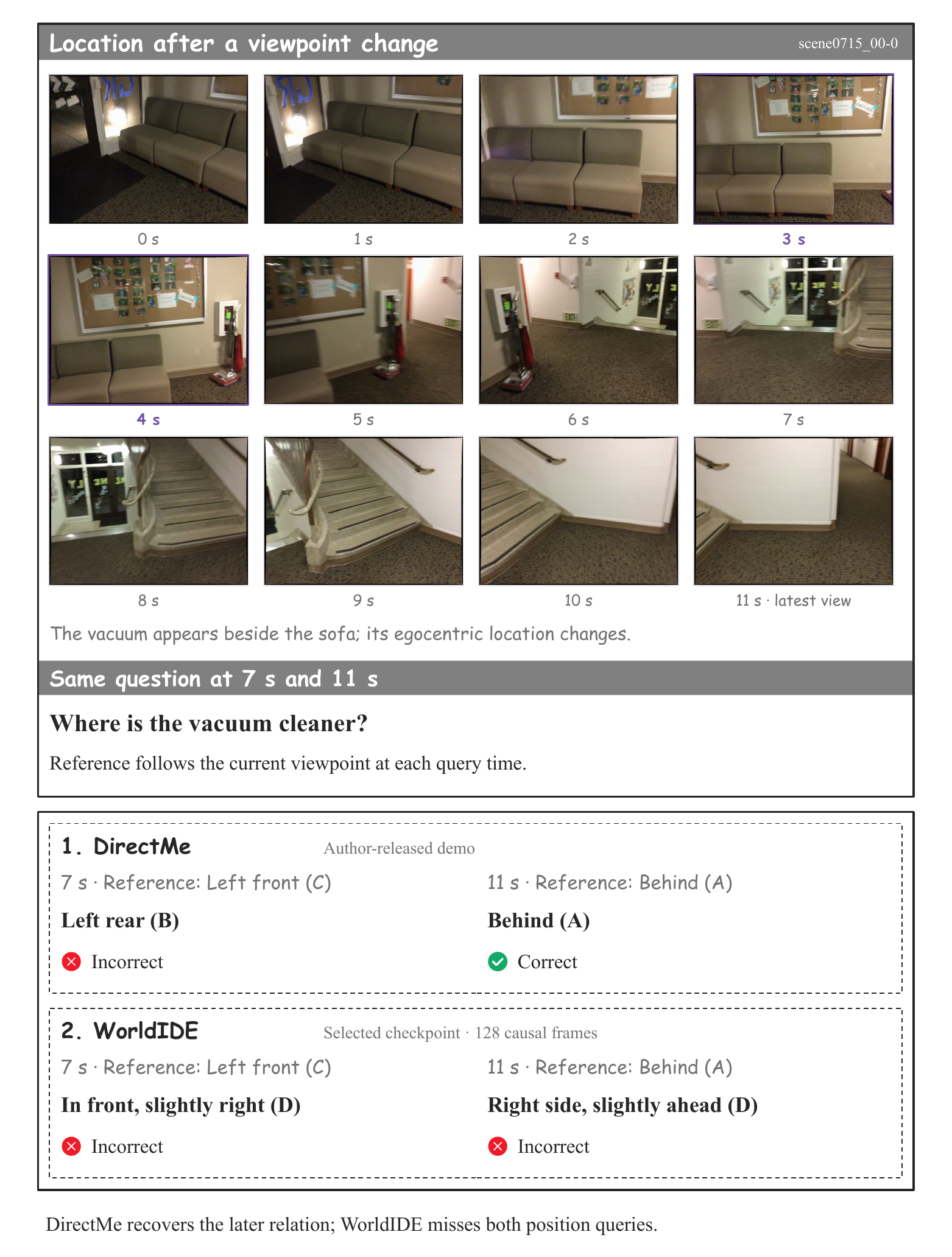}
\caption{\textbf{Egocentric location across a viewpoint change.} In the same video, the annotated vacuum-cleaner relation changes from left front at 7\,s to behind at 11\,s. Both methods fail at 7\,s; DirectMe is correct at 11\,s while WorldIDE remains incorrect. Long answer options are abbreviated in the figure. These are two queries from one scene, not independent videos or a controlled rerun.}
\label{fig:case-directme-position}
\end{figure*}
\clearpage
\section{Supplementary State-Reuse and Readout Diagnostics}
\label{app:case-studies}

These post-hoc cases illustrate source sensitivity and metric recovery using
WorldIDE's native and auxiliary readout protocols. The native case retains
every question for its selected video; population-level results appear in
Appendices~\ref{app:shared-state} and~\ref{app:state-plane}.

\subsection{Native Readout: Six Questions, One State}
\label{app:case-native}

\begin{figure*}[p]
\centering
\includegraphics[width=0.98\textwidth,height=0.79\textheight,keepaspectratio]{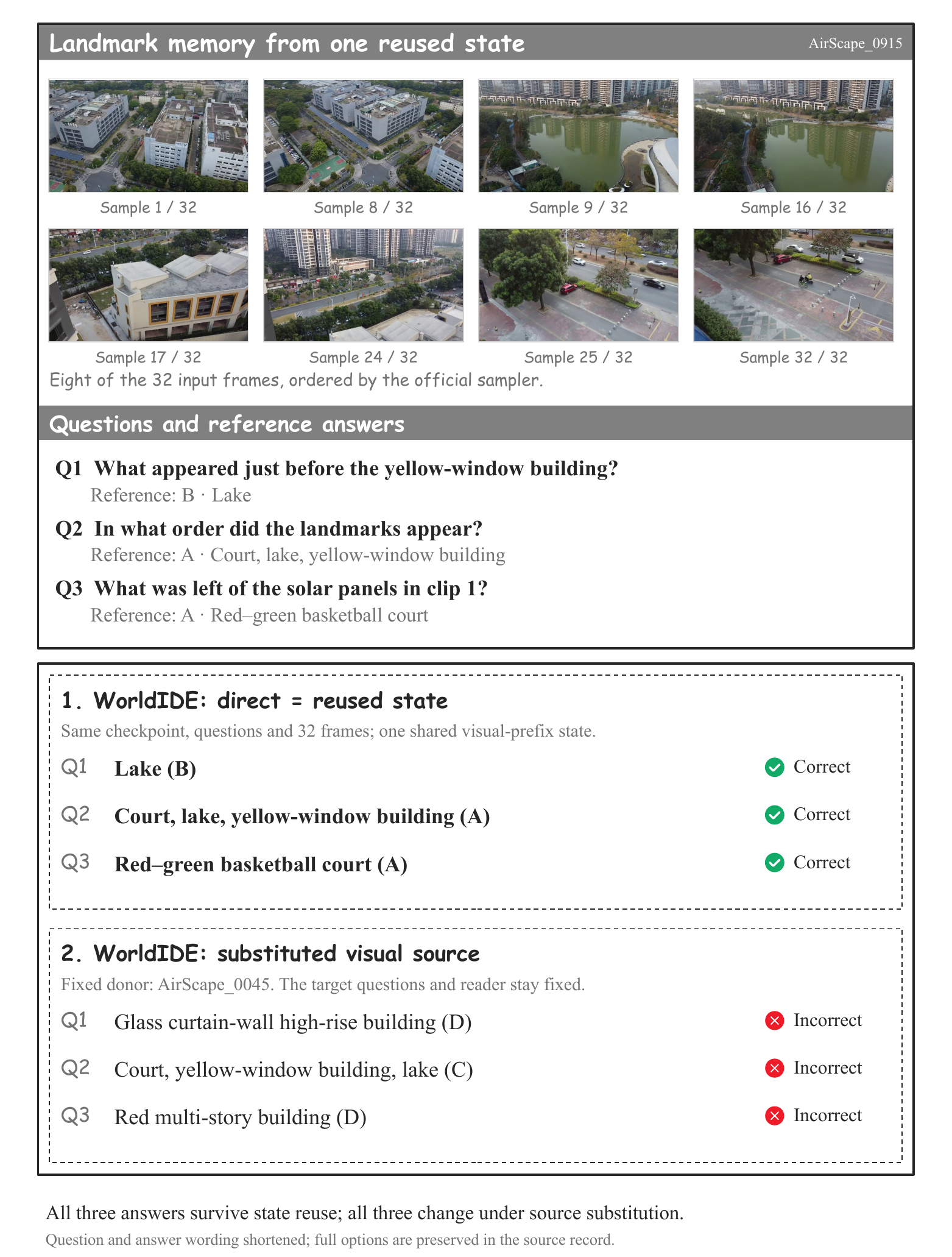}
\caption{\textbf{Landmark memory under visual-state reuse.} Three landmark and position questions from SIS video \texttt{AirScape\_0915}. Direct and shared-prefix inference give the same correct answers. Substituting the prefix state from \texttt{AirScape\_0045} changes all three to incorrect answers. Eight of the 32 input frames are shown. Question and answer wording is shortened.}
\label{fig:case-native}
\end{figure*}

\begin{figure*}[p]
\centering
\includegraphics[width=0.98\textwidth,height=0.79\textheight,keepaspectratio]{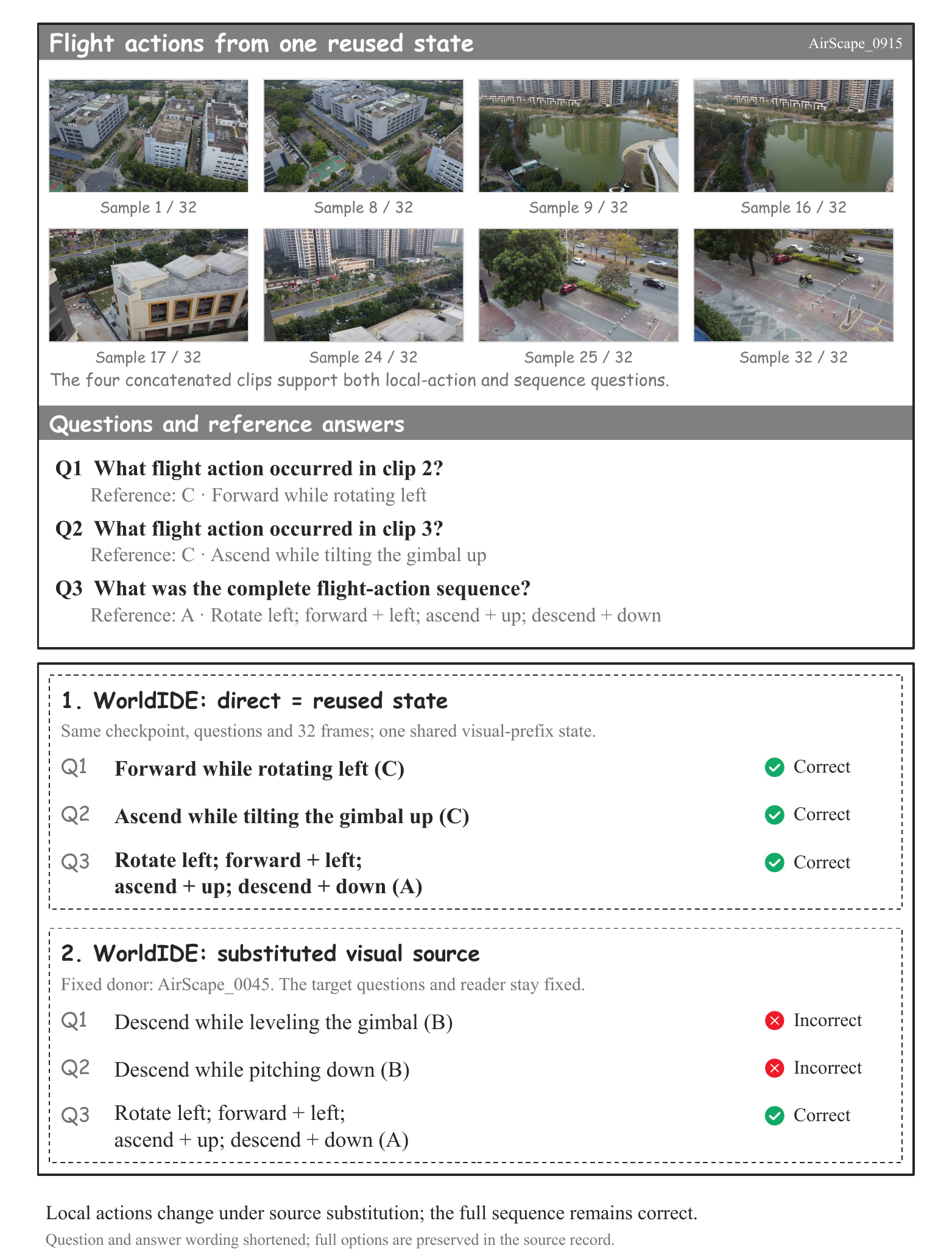}
\caption{\textbf{Flight actions under the same source intervention.} The remaining three questions from \texttt{AirScape\_0915}, completing the six-question case with Figure~\ref{fig:case-native}. Direct and reused-prefix answers agree and are correct. The substituted source changes the two local-action answers, while the complete action sequence remains correct. The sequence is abbreviated in the figure.}
\label{fig:case-motion}
\end{figure*}

Figures~\ref{fig:case-native} and~\ref{fig:case-motion} compare all six SIS questions for
\texttt{AirScape\_0915}. The direct and shared-prefix arms use the same
checkpoint and 32 sampled images, and their recorded option letters agree
exactly on every question. All six shared predictions have the same visual
state identifier. The frozen cyclic intervention substitutes the state from
\texttt{AirScape\_0045}; the target questions, options, decoder, and scorer
remain fixed. Direct and shared inference are both correct on 6/6 questions;
the substituted state is correct on 1/6.

\stitle{Landmark identity and ordering.}
Asked which landmark occurred immediately before the yellow-window-frame
building, both correct-source arms select \emph{the lake} (B), whereas the
wrong-source arm selects \emph{the glass curtain-wall high-rise building}
(D). The error also changes the ordering question: the correct sequence is
red--green basketball court, lake, then yellow-window-frame building (A);
the substituted state places the building before the lake (C). These are
specific changes in retrieved landmark identity and temporal order.

\stitle{Relative position and observed motion.}
For the object left of the ground-level solar panels in clip 1, the answer
changes from \emph{red and green basketball court} (A) to \emph{red
multi-story building} (D). For clip 2, \emph{moving forward while rotating
left} (C) becomes \emph{descending vertically while tilting the gimbal to
horizontal} (B). For clip 3, \emph{ascending while tilting the gimbal up}
(C) becomes \emph{descending while pitching down} (B). The full flight-action
sequence remains A under both sources: rotating left, moving forward while
rotating left, ascending vertically while pitching up, then descending
vertically while pitching down. Thus the intervention changes five concrete
answers while leaving one correct answer intact.

\subsection{Explicit Readout: Geometry, Events, and a Failure}
\label{app:case-plane}

\begin{figure*}[p]
\centering
\includegraphics[width=0.98\textwidth,height=0.79\textheight,keepaspectratio]{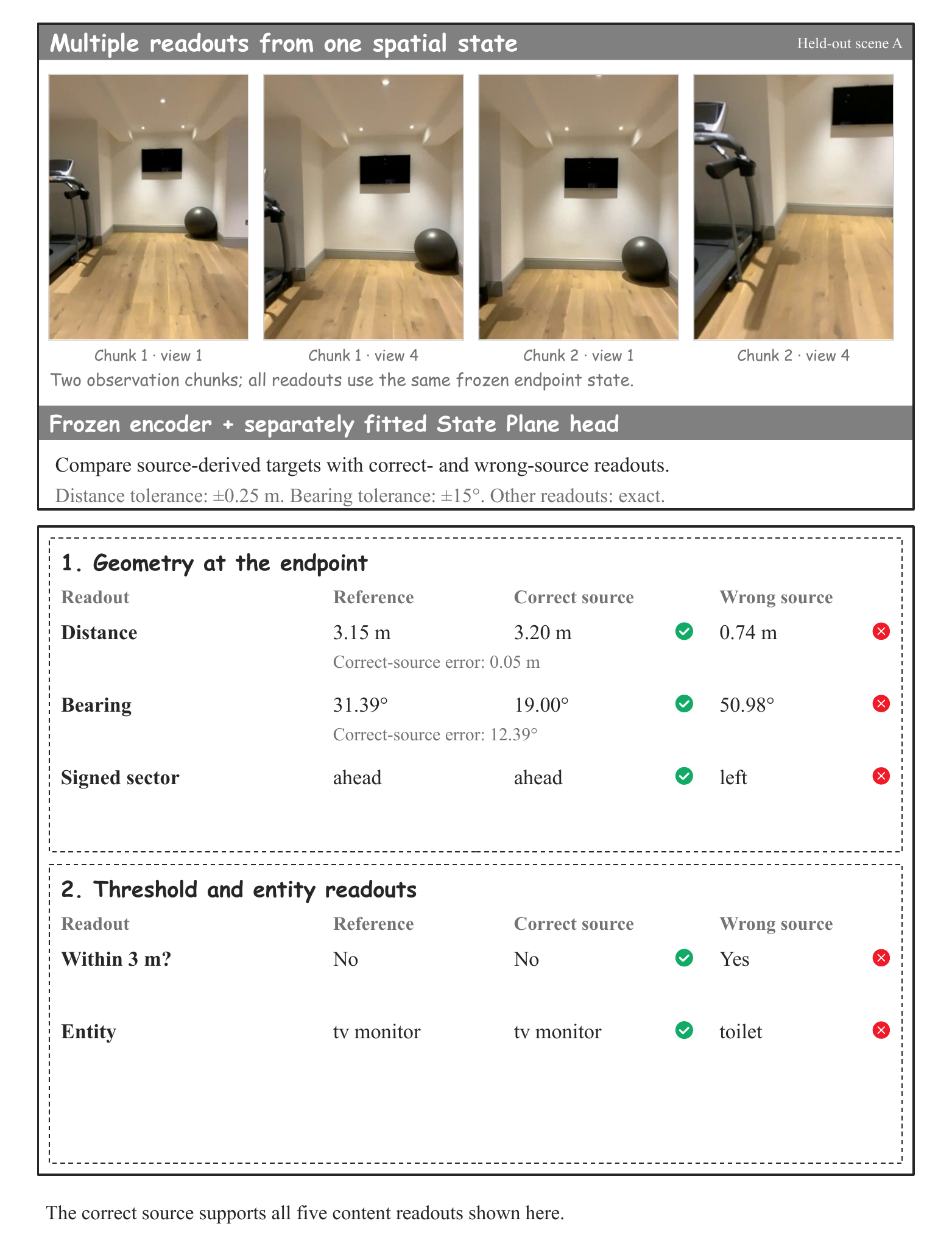}
\caption{\textbf{Multiple content readouts from one auxiliary spatial state.} Four selected views from two observation chunks of held-out scene A. Correct-source predictions pass all five displayed tests, while the substituted source fails them. Distance uses $\pm0.25$\,m tolerance, bearing $\pm15^\circ$, and the other outputs exact equality. Targets come from source geometry.}
\label{fig:case-plane}
\end{figure*}

Figure~\ref{fig:case-plane} uses the auxiliary head fitted to frozen features
of the WorldIDE checkpoint, corresponding to the 63.57\%
correct-source row in Table~\ref{tab:appendix-plane-results}. Each scene's
operators reuse one query-independent endpoint representation. The evaluator
accepts distance error at most $0.25$\,m and circular bearing error at most
$15^\circ$; categorical and event readouts require exact equality. The figures
use the recorded correctness flags and round numbers only for display.

\stitle{Scene A: coupled geometric readouts.}
At endpoint \texttt{d8ea5cb6\ldots::2}, the target entity is a television
monitor. Correct-source distance is $3.1969$\,m against $3.15$\,m, an error
of $0.0469$\,m. Substitution changes the prediction to $0.7413$\,m, an error
of $2.4087$\,m, and flips the ``within 3\,m'' answer from No to Yes. Bearing
changes from $18.9959^\circ$ to $50.9826^\circ$ against a
$31.39^\circ$ target: circular error increases from $12.3941^\circ$ to
$19.5926^\circ$, crossing the scoring tolerance. The derived sector changes
from ahead to left, and the predicted entity changes from television monitor
to toilet. Distance and threshold share a scalar prediction, while bearing
and sector share an angle; the derived decisions reuse these same predicted fields.

\stitle{Scene B: direction and time do not guarantee precise distance.}
At endpoint \texttt{3b8c37a8\ldots::4}, the correct-source state retains
left-of-camera direction and predicts $92.0521^\circ$ against
$101.18^\circ$, within angular tolerance. It also returns chunk 3 for both
the closest and rightmost observations. However, predicted distance is
$1.8312$\,m against $0.98$\,m: the $0.8512$\,m error fails the distance
criterion. The wrong source predicts $1.6696$\,m, $142.8275^\circ$, behind,
chunk 2 for closest, and chunk 1 for rightmost. This example separates
successful directional/event access from fine metric accuracy. The turn index
is fixed by the protocol and is omitted from the illustrated evidence of
learned content.

\begin{figure*}[p]
\centering
\includegraphics[width=0.98\textwidth,height=0.79\textheight,keepaspectratio]{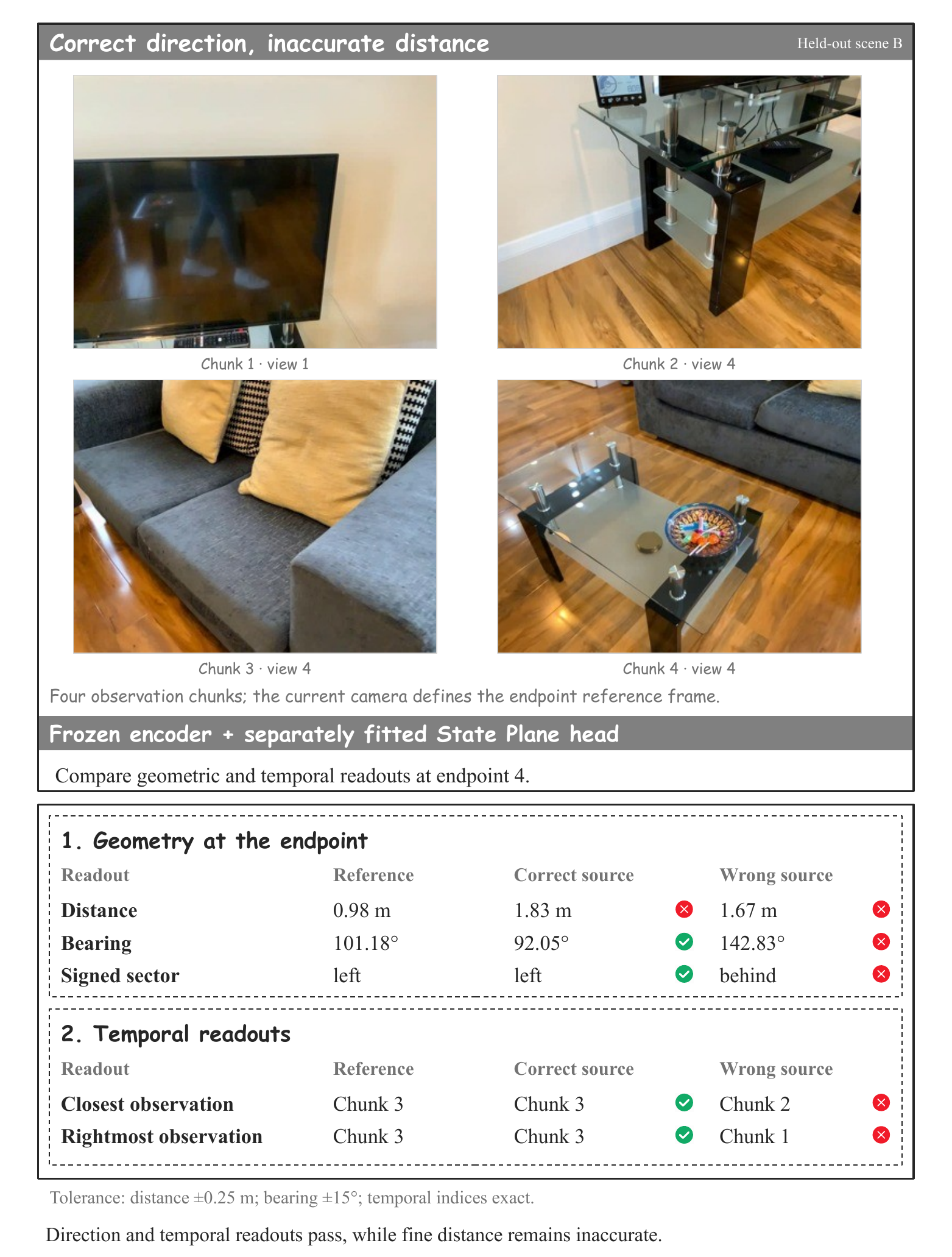}
\caption{\textbf{A fine-distance failure despite correct direction and event readouts.} Four selected views from four observation chunks of held-out scene B. The correct-source distance prediction is 1.83\,m against a 0.98\,m target, outside the $\pm0.25$\,m tolerance. Bearing, sector, closest observation and rightmost observation pass their recorded tests. The wrong-source predictions are retained alongside them. This selected failure complements scene A in Figure~\ref{fig:case-plane}.}
\label{fig:case-metric-failure}
\end{figure*}
\clearpage

\section{Additional Related Work}
\label{app:related}

\stitle{Spatial reasoning in vision-language models.}
Visual instruction tuning \citep{NEURIPS2023_6dcf277e} made answer
supervision the standard route to new multimodal skills, and spatial
variants generate question--answer pairs from 3D or depth annotations
\citep{chen2024spatialvlm,cheng2024spatialrgpt}. Diagnostic benchmarks show
that such models still confuse basic relations such as left and right
\citep{kamath-etal-2023-whats}, fail perception tasks that humans solve at a
glance \citep{Fu_2024}, and lag on embodied spatial questions
\citep{du-etal-2024-embspatial}; VSI-Bench extends the diagnosis to how
models see, remember, and recall spaces from video \citep{Yang_2025_CVPR}.
Training remedies include simulated dynamic spatial tasks \citep{ray2025sat}
and dedicated 3D encoders for video spatial intelligence
\citep{wu2025spatialmllm}. Prior work establishes the value of spatially grounded supervision for
visual reasoning. \method{} organizes complementary spatial and temporal
relations across the training distribution and examines their effects through
cross-history distinguishability and functional spatial readout.

\stitle{Streaming video and reusable visual state.}
Online video LLMs interleave perception and response over streams
\citep{Chen_2024_CVPR}, compress clips into memories that later questions
consult \citep{NEURIPS2024_d7ce06e9}, separate perception from decision and
reaction for real-time interaction \citep{Qian_2025_CVPR}, or retrieve cached
video key--value states for streaming question answering
\citep{di2025streaming}. In medical VLMs, MedEyes selects image regions
progressively \citep{zhu2026medeyes}, and MedSynapse-V builds latent
diagnostic memory \citep{zhu2026medsynapse}.
On the serving side, paged and radix-tree KV
caches and modular prompt caches reuse shared prefixes across requests
\citep{Kwon_2023,NEURIPS2024_724be447,MLSYS2024_a66caa17}.
\method{} uses the
same mechanism as an experimental object: the shared visual prefix $Z_h$ is
the state whose reuse and source dependence RQ2 measures.

\stitle{Probing for world state.}
Linear probes read information from frozen intermediate representations
\citep{alain2018understandingintermediatelayersusing}, with control tasks
and probe-capacity analyses separating what a representation encodes from
what a probe learns \citep{hewitt-liang-2019-designing,belinkov-2022-probing}.
Probes have recovered entity and world states from language models
\citep{li-etal-2021-implicit,kim-schuster-2023-entity}, board states from a
sequence model trained on game moves \citep{li2023emergent}, and linear
representations of space and time \citep{gurnee2024language}. The State
Plane follows this methodology with matched, separately fitted heads on
frozen encoders and deterministic operations over the recovered typed fields.

\stitle{World models, auxiliary tasks, and scene memory.}
Learned world models compress observations into a state used for prediction
and control \citep{NEURIPS2018_2de5d166}, and auxiliary prediction tasks are
a long-standing way to shape such representations
\citep{jaderberg2016reinforcementlearningunsupervisedauxiliary}. Embodied
agents also maintain explicit 3D scene graphs or scene memories for
planning \citep{Gu_2024,Yang_2025_CVPR_3DMem}. \method{} applies relational supervision to visual histories and examines
the resulting representations through native language readout and matched
spatial probes.